# A WORD GRAMMAR OF TURKISH WITH MORPHOPHONEMIC RULES

A THESIS SUBMITTED TO
THE GRADUATE SCHOOL OF NATURAL AND APPLIED SCIENCES
OF
THE MIDDLE EAST TECHNICAL UNIVERSITY

BY

SERDAR MURAT ÖZTANER

IN PARTIAL FULFILLMENT OF THE REQUIREMENTS FOR THE

DEGREE OF

MASTER OF SCIENCE

IN

THE DEPARTMENT OF COMPUTER ENGINEERING

JANUARY 1996

Approval of the Graduate School of Natural and Applied Sciences.

Prof. Dr. İsmail Tosun
Director

I certify that this thesis satisfies all the requirements as a thesis for the degree of Master of Science.

Prof. Dr. Neşe Yalabık
Head of Department

This is to certify that we have read this thesis and that in our opinion it is fully adequate, in scope and quality, as a thesis for the degree of Master of Science.

Assist. Prof. Dr. Cem Bozşahin
Supervisor

Examining Committee Members

Prof. Dr. İbrahim Akman

Assist. Prof. Dr. Cem Bozşahin

Assist. Prof. Dr. Faruk Polat

Assist. Prof. Dr. Göktürk Üçoluk

Assoc. Prof. Dr. Deniz Zeyrek

# ABSTRACT

## A WORD GRAMMAR OF TURKISH WITH MORPHOPHONEMIC RULES

Öztaner, Serdar Murat

M.S., Department of Computer Engineering

Supervisor: Assist. Prof. Dr. Cem Bozşahin

This thesis is about the computational morphological analysis and generation of Turkish word forms. Turkish morphological description is encoded using the two-level morphological model. This description consists of a phonological component that contains the two-level morphophonemic rules, and a lexicon component which lists lexical items (indivisible words and affixes) and encodes the morphotactic constraints. In the scope of the study, a generic word grammar in a tabular form expressing the ordering relationships among morphemes is designed and morphophonemic processes along with solutions to exceptional cases are formulated.

# ÖZ

## TÜRKÇE'NİN SÖZCÜK YAPISI VE BİÇİMBİRİM SESBİRİM ETKİLEŞİMİ İÇİN KURALLAR

Öztaner, Serdar Murat

Yüksek Lisans, Bilgisayar Mühendisliği Bölümü

Tez Yöneticisi: Yrd. Doç. Dr. Cem Bozşahin

Bu tez Türkçe sözcüklerin bilgisayar ortamında biçimbilimsel çözümlemesi ve üretimi ile ilgili bir çalışmadır. Bunun için Türkçe'nin biçimbilimsel tanımlaması iki düzeyli biçimbilimsel model kullanılarak kodlanmıştır. Bu tanımlama iki öğeden oluşur: Biçimbirim değişmelerine yol açan ses olaylarını modelleyen iki seviyeli kuralları kapsayan sesbilimsel öğe, ve kökler ile ekler arasındaki sıralama ilişkilerini formüle eden biçimdizimsel öğe. Çalışma kapsamında, Türkçe için genel sözcük yapıbilgisi tablolar halinde hazırlanmış ve biçimbirim değişmeleri sıradışı durumları kapsar şekilde kurallanmıştır.

To my family. Önüt Öztaner, Nesrin Öztaner, Burak Öztaner, Şebnem Öztaner.

# ACKNOWLEDGMENTS

In the first place, I am thankful to all the research assistants of METU Computer Engineering Department for not only their friendship but also their comments that motivated me more and more to work harder for a better implementation. I want to thank our dear system administrators Onur Tolga Şehitoğlu and Mehmet Uğur Yılmaz who are always ready to solve any problem. I am very much indebted to Sema Mançuhan who helped me in programming the user interface. Special thanks go to Aylin Akça and Arzu Şişman for their support throughout this thesis.

Many thanks to all the members of the *LcsL* group who feel, face, and cope with the same things. I especially owe a lot to Elvan Göçmen, Onur Tolga Şehitoğlu, and Tolga Yıldırım who were always willing to listen to the problems and to provide me with the solutions. I owe a special word of gratitude to Tolga Yıldırım; if he were not there, this work would not be completed in time. His assistance will never be forgotten.

Thanks to NATO (project name TU-LANGUAGE) and TÜBİTAK (project no. EEEAG-90) for supporting this project. Without the software and hardware they have supplied, this work would not be possible.

I am thankful to Assist. Prof. Dr. Cem Bozşahin for not only his supervision but his understanding in spite of the repetitive mistakes I made in the manuscripts of this thesis. I also owe thanks to Assoc. Prof. Dr. Deniz Zeyrek for her precious comments about the framework.

Finally, my family gave me the greatest support. I am and will always be grateful to them.

# TABLE OF CONTENTS

# LIST OF TABLES

# LIST OF FIGURES

# CHAPTER I

## INTRODUCTION

Morphological analysis is the process of breaking up a word into its components, called the root and the affixes. In doing so, morphological analyzers determine the grammatical (e.g., person, number, case) and morphological (i.e., derivational) properties of the word.

In an all-suffixing language like Turkish, morphological analysis determines the root, the suffixes, and their categories. For instance,

(1) a. çiçek -çi -de -ki -ler -in -ki
flower DER(house) CASE REL PLU GEN REL
'the one that belongs to those who are in the flower shop'

b. gör -üş -tür -t -tür -ül -e -me -yebil -ir -di
see RECP CAUS-T CAUS-A CAUS-I PASS ABIL NEG POSS TENS AUX
'it could not have been made possible for he/she to have someone to get someone else to meet.'

Affixes can be realized in many different ways, e.g., /-ler/-lar/, /-cı/-ci/-cu/-cü /-çı/-çi/-çu/-çü/. These alternations are not due to ordering constraints; they are conditioned by the phonological properties of the surronding sound segment of the affixes. However, the so-called morphotactical constraints which involves the ordering relations among morphemes are also effective in realization of word forms. That is, affixes can not be attached one after another in a free order. For instance, the plural suffix -lAr strictly precedes the case suffixes in a word form. Examples below illustrate such ordering constraints of of the nominal paradigm.[1]

```
(2) a. araba -lar  -dan        *araba -dan   -lar
       car   -PLU -CASE        *car   -CASE -PLU
       'from the cars'

    b. ev    -imiz  -de         *ev    -de    -imiz
       house -POSS -CASE        *house -CASE -POSS
       'at our house'
```

Thus in designing a morphological analyzer, one should design and implement the mechanisms that performs the phonological alternations and checks the validity of ordering for the realization of morphemes. Many theoretical computational models have been proposed for natural language processing applications, especially for morphological analysis. Some tools have been developed on these models. A major problem with most of these tools is that they are language-dependent. Among the models proposed, the *two-level* model provides a language independent model for recognition and production of word forms. The model is bidirectional, that is, it is capable of both analyzing and synthesizing word-forms. Two-level model has been implemented as operational computer programs on various systems.

This study is about the design and implementation of a morphological processor for Turkish. It is capable of analyzing and synthesizing word forms. A commercial system based on two-level model has been used. This involves the implementation of morphophonemic processes as two-level rules and morphotactical constraints using finite-state mechanisms. A generic word-grammar in a tabular form which can serve the basis for some other possible implementations of Turkish morphology on different systems have also been provided. The mechanisms for the integration of the morphological module with the lexical and syntactical components are discussed.

[1] Examples marked with '*' are considered ungrammatical or non-felicitous.

This morphological analyzer can be used as a part of natural language processing applications on Turkish. It is to be integrated to an MT system to be developed in the scope of the TU-Lang Project.

# CHAPTER II

## BACKGROUND

Words have phonological and morphological properties. Phonological properties determine their surface form, and morphological properties specify their meaning-bearing constituents. In this chapter, we provide a short summary of related concepts in phonology and morphology.

## II.1 Phonology

### Phonetics

*Phonetics* is the study of the inventory of all speech sounds which humans are capable of producing. The term "speech sound" arises from the fact that not all noises which we are capable of producing with our vocal apparatus are employed in speech (i.e., coughing, sneezing, snoring etc.). *Articulatory phonetics*, the study of speech production, is the branch of phonetics on which most phonological theories are based.

Consonants are produced by obstructing in some way the flow of air through the vocal tract. The production of consonants involves four major parameters which can be varied independently of each other to create different kinds of consonants. The four parameters are [16]:

a) The *airstream mechanism*, which refers to the way in which moving body of air that provides the power for speech production is generated.

b) The *state of the glottis*, which classifies sounds as being voiceless and voiced. *Voiceless* sounds are produced when there is a wide open glottis, with a big space between the vocal cords; *voiced* sounds are produced when the vocal cords are close together so that the air has to force its way through them.

c) The *place of articulation* refers to the place in the vocal tract where the airstream is obstructed in the production of a consonant (e.g., bilabial, alveolar, palatal).

d) The *manner of articulation* refers to the way in which the airstream is interfered with in producing a consonant (e.g., plosives, fricatives).

The consonants of Turkish are given in Subsection IV.1.1 along with their features (see Table IV.2).

Vowels are more difficult to describe accurately than consonants. This is largely because there is no noticeable obstruction in the vocal tract during their production. Vowels are typically voiced, but they have no place or manner of articulation. Therefore, a different set of concepts has been used for the description of vowels.

For instance, depending on the location of the highest point of the tongue,

vowels are regarded as *front*, *central* or *back* [1]. The second criterion is the shape of the lips. Vowels are classified as being *round* and *unrounded* [2]. The third factor determines the characteristic of being *high* (close) or *low* (open), according to the amount of space left between the palate and the tongue. All the vowels which have been described so far are *monophthongs*, i.e., vowels whose quality remains virtually unchanged throughout their duration. In addition to such vowels, some languages (English included) also have *diphthongs*[3] i.e., vowels whose quality changes during their production.

**Phonemes**

Speech sounds can be written without reference to articulatory details. *Phoneme* is the abstraction of sounds that actually occur. Members of the same phoneme family, i.e., the various physically distinct sounds which count as variations of a given phoneme are called *allophones.*

A phoneme is conventionally represented by a letter symbol between slant lines. For instance, two diverse voiceless stops in English are grouped into two phonemes labelled /t/ and /k/, each of which has a range of allophones which differ slightly from each other[4]. When you pronounce the words `car` and `key` and observe the movement of your tongue, you can feel that the “k” in `car` is made farther back than that in `key`. Although these two varieties of “k” are physically different, they are not functionally different in English. This distinction cannot be used to contrast the word meaning in English.

---

[1] Also called as the feature of *Gravity* [24]
[2] Also called as the feature of *Flatness* [14]
[3] For example, `pie, buy, cry`.
[4] The examples given are from [16].

### Distinctive features

What is really happening in phonology is happenning in terms of seperate phonetic properties, not of "unitary sounds". It is these phonetic properties rather than phonemes that are basic building blocks of phonology. These phonetic and phonological properties are called the *distinctive features*[5]. Actually, languages construct their individual phoneme systems by selecting different combinations from this small inventory of phonetic properties.

Current distinctive feature theory has its roots in the work of Trubetzkoy and Jakobson in 1939. Jakobson, Fant and Halle [14] concentrated their investigations on phonological oppositions that occur universally and proposed a set of such features. Robert B. Lees [24] presented Jakobsonian features for Turkish phonemes in his work *The phonology of modern standard Turkish.* Because of some shortcomings of the Jakobsonian features, Chomsky and Halle [10] proposed a major revision of the theory of distinctive features. This new set of features is known as the SPE system. The selected features for Turkish will be mentioned when the morphophonological model is introduced.

### Phonetic vs phonemic transcription

Since standard ortography fails to provide an adequate representation of speech, a proper accurate way for this representation has to be provided. There may be at least two distinct levels of representation: The phonetic level and the phonemic level. The representation used in phonetic level is the *phonetic transcription*

[5] The *distinctive features* are the basic phonological ingredients which phonemes are made of. They characterise the phonological contrasts like ±*back*, ±*high*, ±*round* etc. found in all the world's languages.

(also called broad transcription) which only shows functional differences, i.e., differences between sounds which are used to distinguish word meaning. It only uses enough symbols to represent each phoneme of the language in question with a symbol of its own. *Phonetic transcription* (also called narrow transcription) on the other hand, is much more detailed and attempts to provide a more faithful representation of speech. The phonemic (broad) transcriptions for the words *attend* and *two* are /əttend/ and /tu/ respectively. The phonetic (narrow) descriptions for the same words are [ə$t^h$end] and [$t^w$u][6].

One thing that needs to be borne in mind throughout this work is that computational analysis of morphology assume *written* input. Therefore, phonemic and phonetic transcriptions is used at the minimum, only when there is a need to formulate phonological rules of the language.

### Phonological processes

As stated above, a phoneme may have several allophones. The allophone selected in a particular position is dependent on the other sound segments that are adjacent to it. The most common phonological process is *assimilation*. Assimilation is the modification of a sound in order to make it more similar to some other sound in its neighbourhood. The advantage of having assimilation is that it results in smoother, more effortless, more economical transitions from one sound to another. The speaker usually tries to conserve energy by using no more effort than is necessary to produce an utterance. The alternation in the phonological realizations can be accounted for in terms of assimilation. A typical assimilation process in Turkish is the vowel harmony where a preceeding vowel affects the succeeding ones. There are various common assimilation processes found in the languages of the world, such as *palatalization*, *labialization*, *nasalization* etc. Besides assimilation, there are processes which add, delete, or duplicate phonological material.

[6] A *diacritic* which is a part of the phonetic inscription, is a symbol that indicates a phonetic feature. For instance $_+$ indicates backness, $^w$ indicates labialization, and $^h$ indicates aspiration.

## II.2 Morphology

Morphology is the branch of linguistics that deals with the word structure and how words are formed.

To gain a better initial understanding of morphology, it may help observing the morphological properties of a selected form (e.g., `ekmeği` from Turkish).

- `ekmeği` is the inflected form of the word `ekmek`.
- The word `ekmeği` is formed by appending the accusative case suffix, `-(y)H`, to the stem `ekmek` (i.e., `ekmek`+Case) or the third person singular suffix, `-(s)H`, (i.e., `ekmek`+Possesive) to the stem.
- In the form `ekmeği`, the stem final consonant has changed from /`k`/ to /`ğ`/.

The three points above touch on a broad range of issues in morphology. The first point brings up the question of the function of morphology (i.e., what kinds of information morphological forms convey). In this particular example, as an inflection of the word `ekmek`, the form `ekmeği` means both "somebody's bread" and "bread-ACC". While the first meaning imputes the role of being the possessed noun of a genitive noun group to the form `ekmeği`, the latter meaning marks the form `ekmeği` as a definite object. The second point raises the issue of morphologically complex words and out of what kinds of "pieces" they are built. Finally, the third point brings out the question of how rules of phonology (orthography) interact with morphology. These observations raise a number of issues [33]:

a. How are words built up from smaller meaningful units (i.e., *morphemes*)? What are morphemes, and what can they look like?

b. What kinds of things can morphology do in different languages? What are the functions of morphology? These issues are related with the range of information conveyed by morphology.

c. What are the constraints on the order of morphemes in words?

d. How do the phonological rules affect the morphological analysis?

## II.3 Morphological processes

It has been mentioned that the words are formed by combination of smaller meaningful units. In that respect both roots and affixes[7] are morphemes. The main difference between roots and affixes is that affixes must attach to some stem and usually specify requirements on the properties that must be present on the stem. Roots (and stems), in contrast, do not specify attachment requirements in the sense that thay are the things to which affixes attach. Thus, while roots (at least in many languages) can be considered to be free (*unbound*) morphemes as they may be used as words without any further morphological attachment, affixes by definition must always be *bound* morphemes.

[7] A segment of phonemes added to the beginning of a word (in the case of *prefix*) or to the end of a word (in the case of *suffix*) to change its meaning or its use. There are also *infixes* and *circumfixes*.

The simplest model of morphology is the case where a morphologically complex word can be analyzed as a series of morphemes concatenated together. An example from English is `antidisestablishmentarianism`. Such morphology is typically called *concatenative morphology* where morphemes are strung together like "beads on a string". Agglutinative languages are defined as languages whose morphology, especialy the inflectional one, is completely concatenative, and where fairly long words can be formed by stringing together of morphemes. Turkish is such a language where suffixation is almost the only morphological device. Although Turkish is an exclusively suffixing language, there are, however, a few very unproductive prefixes[8] of foreign origin, such as `na-` ("un-" in English). A typical example from Turkish is illustrated in Figure II.1. The layout of this example is as follows:

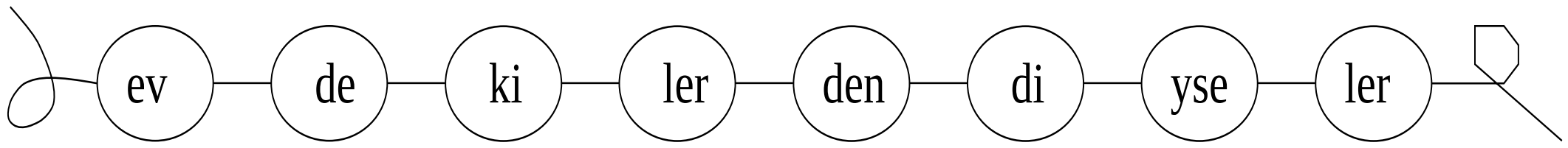

Figure II.1: An example of purely concatenative morphology

```
(3)   ev    -de  -ki  -ler -den -di   -yse  -ler
      house -LOC -REL -PLU -ABL -PAST -COND -PERS
      'if they were from those at home'
```

Languages can of course have a concatenative morphology which allows prefixes: the English example `antidisestablishmentarianism` has two prefixes, `anti-` and `dis-`. However, there is an asymetry between prefixation and suffixation cross-linguistically in that suffixing is far more common than prefixing.

[8] Besides loanwords, the only prefixation in Turkish is the intensifier: `çarçabuk`, `besbelli`, `kıpkırmızı`, etc. But this is a phonological process, not a true suffix.

Concatenative morphology is a convenient kind of morphology from a computational point of view, since one can often write simple finite-state grammars which can adequately describe such morphology.

Morphemes can combine to form words in other ways apart from simple linear concatenation (i.e., *prefixation* and *suffixation*). These are *nonconcatenative processes* covering a spectrum of phonological and morphological phenomena that require special treatment, including *infixation*, *circumfixation*, *reduplication*, *gemination*, *degemination*, and *metathesis*. In the case of infixation, an infix is a morpheme similar to a prefix or suffix; it differs in that it is inserted into the middle of another morpheme rather than simply concatenated to the beginning or end. In the case of *circumfixation* affixes attach discontinuously around a stem. In the case of *reduplication*, part of phonological material of a word is copied and attached to the word. *Gemination* is the doubling of a phonological segment; like reduplication, it involves the process of copying. *Degemination* is the reduction of a sequence of two identical segments to a single segment. *Metathesis* is a process that interchanges the linear sequence of two segments. Table II.1 shows examples[9] of various morphophonological process types.

Table II.1: A sample group of morphophonological processes

| process | language | | | | | |
|---|---|---|---|---|---|---|
| `infixation` | `Philip.` | `fikas` | "strong" | → | `f+um+ikas` | "be strong" |
| `circumfixation` | `German` | `haben` | "have" | → | `ge+hab+t` | "had" |
| `reduplication` | `Tagalog` | `pili` | | → | `pipili` | |
| `gemination` | `Hebrew` | `melek` | "king" | → | `hammelek` | "the king" |

Note that all the morphological processes above involve the interaction of phonological material. Thus, the concept of morpheme has to be redefined to include any kind of phonological expression of a morphological category that involves the addition of phonological material.

[9] The examples are from [33] and [1].

In all the forms of morphology presented above, affixes (more generally morphemes) consist phonologically of a sequence of segments. Yet, there is another model of morphology, the so-called *root-and-pattern morphology* of Arabic where morphemes are nonsegmental.

The rules that constrain the ways in which morphemes are put together are called *morphotactics*. In languages like Turkish and Finnish, which have concatenative models of morphology, morphotactics are nothing but ordering restrictions on morphemes. Morphotactics of Turkish will be discussed in Section V.2.

The interpretation of a sentence requires information about the syntactic and semantic functions of the various words in the sentence. This information may be provided by morphology. From the point of view of someone building a morphological analyzer, it is important to bear in mind the range of information that may be conveyed by morphology. How much information and what kind of information is conveyed morphologically?

It will be better to answer this question by giving examples from Turkish nominal morphology. Before that, let us introduce the concepts of *inflectional* and *derivational* morphology, since these two are different means of creating word forms. Inflectional morphology takes as input a word and outputs a form of the same word appropriate in a particular context. On the other hand, derivational morphology is supposed to take as input a word and output a different word that is derived from the input word.

When a word is inflected, its part of speech does not change, but the inflected form usually marks a certain inflectional category. For instance in Turkish,

```
(4)    ev-de
       home-LOC
       'at home'
```

is a nominal inflection. The part of speech of `ev` does not change when inflected by the locative suffix `-de` (i.e., `ev` is noun just as `evde`). However, the syntactic role of `ev` differs from that of `evde`: `ev` is a noun in nominative case and therefore, may be the subject or the direct object of a sentence, whereas `evde` is a noun in locative case and therefore, acts as an adverbial phrase expressing the location. Thus, inflection is often required by the syntax.

Derivation may change the part of speech of the word but has nothing to do with the syntax. When a new word form is derived from a word, it acquires a new meaning (often relevant to that of original). For instance in Turkish,

```
(5) a. şeker-lik
       sugar-DER
       'sugar container'

    b. şeker-li
       sugar-DER
       'sweet'
```

are two derivations. In the first case, a new noun is derived from a noun and in the latter, an adjective is derived from a noun. From the syntactic point of view, both word forms are nominals subject to the same inflectional paradigm. There are two issues of derivational morphology that should be discussed specifically for Turkish: productivity and lexicalization. Productivity determines whether a suffix is productive (i.e., applicable to a large set of stems) in deriving new word forms. Lexicalization is about to which extent and in what conditions the derived word forms are no more accepted as derivational elements but distinct stems that should be stored in the lexicon.

**Phoneme-Morpheme interaction: Morphophonemics**

Ideally, the task of a morphological analysis system would be to break the word down into its component morphemes, and, for some applications, determine what the resulting decomposition were to mean. Things are not that simple, however, because of the effects of phonological processes such as assimilation; morphemes often change their shape. The alternative realizations of a morpheme are called *allomorph*s. For instance, in Turkish, the plural suffix `-lAr` have two allomorphs: `-ler` and `-lar`.

## II.4 Computational morphology

In the computational field, there is a fairly broad range of technological applications which make use of morphological information. The computational areas where morphological information have been used in various ways can be listed as follows:

- natural language applications (parsing, text generation, machine translation, dictionary tools and lemmatization)
- speech applications (text-to-speech systems, speech recognition)
- word processing applications (spelling checkers, text input)
- document retrieval

Apart from the applications above, computational morphology has also been used in developing teaching tools for studying morphology itself, both for language learners [13] and for students of morphological theory [22].

Morphological information can be used for the purposes of either analysis or generation. Morphological analysis (parsing and recognition) is concerned with retrieving structure, the syntactic and the morphological properties, or the meaning of a morphologically complex word. Morphological generation is concerned with producing an appropriate morphologically complex word from a set of semantic or syntactic features [33].

Traditionally, computational linguistics has concentrated directly on syntax. The basic reason for this is that the economically dominant languages of the world have relatively simple morphologies. Implementors of natural language processing systems for these languages have been able to handle the morphology by storing all inflected forms in the lexicon, or by using pattern-matching techniques. Morphophonemic alternation, if adressed at all, has been coded as ad hoc procedures directly in the computer program. These methods were designed to work for only the language of interest. No claims of general theoretical interest were made. A nontrivial morphological analysis of any natural language is a necessity if one needs to handle the words of that language in any general way using computational means. Otherwise one needs to have a huge lexicon where all the words along with all their inflectional and derivational forms are listed.

Although some studies on morphological decomposition using computational methods date back to the early 1960s (DECOMP system of MITalk project), computational morphology began to emerge as a discipline in its own right about fifteen years ago. The two-level morphological model, a language-independent model by Kimmo Koskenniemi [19, 20], has served as a convenient starting point for much of the other work on computational morphology.

In this study, Turkish morphology is modelled using the two-level approach. The two-level morphological model will be discussed in detail and compared with other approaches to computational morphology in the following chapter.

# CHAPTER III

## TWO-LEVEL MORPHOLOGY

The two-level model provides a general language-independent mechanism that can be used in morphological analysis of natural languages for language-specific descriptions. The structure of language-specific description is based on the linguistic distinction between phonology and morphology. Therefore, two-level model comprises two components:

- a "rules" component, which contains phonological rules (actually morphophonemic rules). These rules describe the phonologically conditioned changes in morphological forms. Phonological rules are implemented by a set of *finite state transducers.*

- a "lexicon" component, or lexicon, which lists lexical items (indivisible words and morphemes) in their underlying forms, and encodes morphotactic constraints. Morphotactics are implemented by a set of *finite state automata.*

The significance of the two-level model is the theoretical model of phonology used (called *two-level* phonology), in which FSTs operate in parallel. Indeed, the term *two-level morphology*, used by Koskenniemi to describe the framework, is really a misnomer: the "two-level" part of the model describes the method for implementing phonological rules, and has no direct relation with morphology. In the two-level approach, phonology is treated as the correspondence between the *lexical level* of underlying representation of word forms and their realization on the *surface level*. The model is bidirectional and is capable of both analyzing (i.e., recognizing) and synthesizing (i.e., generating) word forms. An abstract illustration of how two-level model functions, can be seen in Figure III.1.

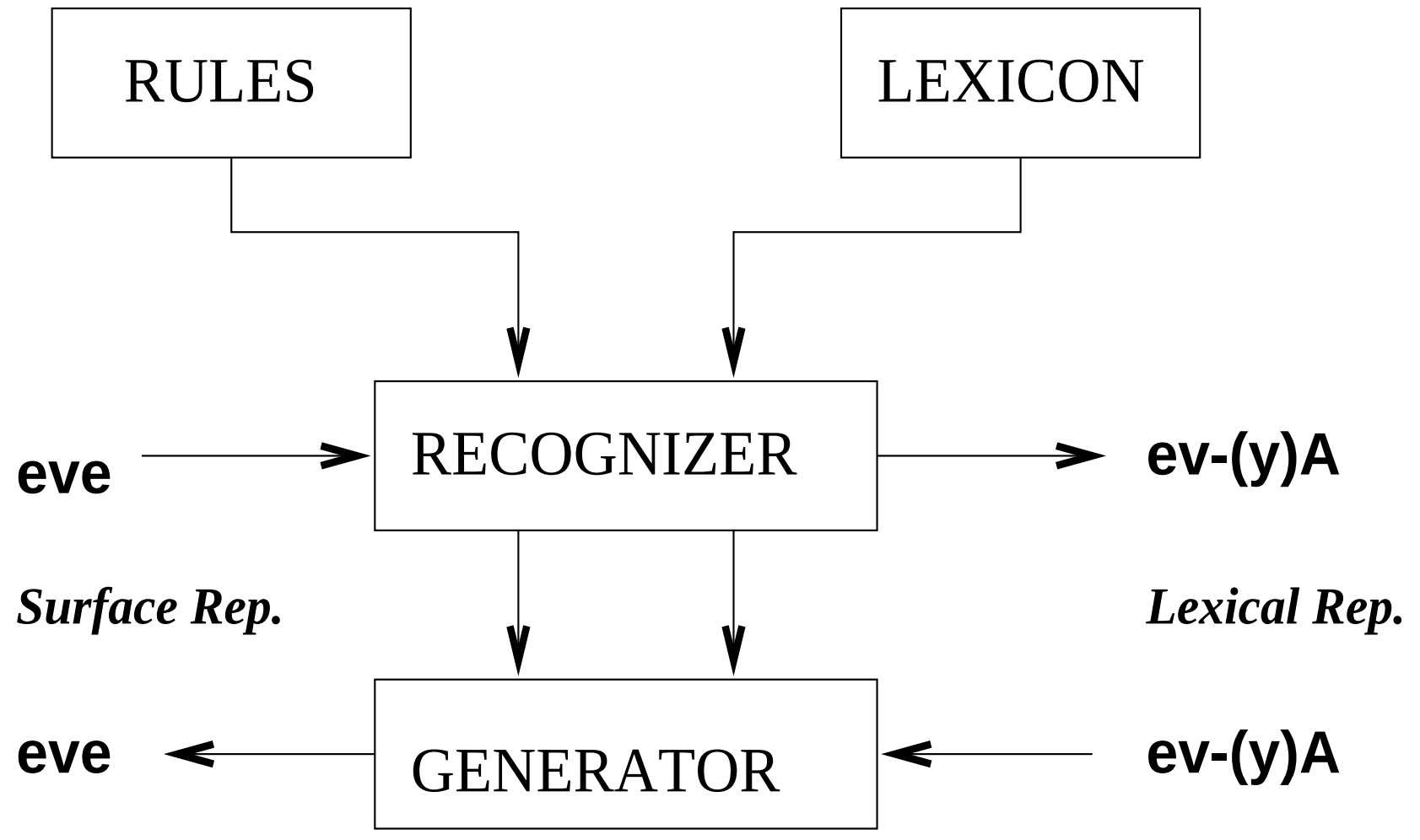

Figure III.1: An overview of how two-level model functions

### Finite-state automata

The simplest finite state machine is a *finite state automaton* (FSA), which recognizes (or generates) the well-formed strings of a *regular language*. The Figure III.2 shows a FSA that defines the regular language $L_1$=$\{ab^na\}$ where $n \geq 0$.

Let me express why and how the finite state automaton is used to model the morphotactics of a language. For the time being, make the false assumption that concatenation is the only way in which morphemes can be combined to form words in a language. Clearly the simplest way to build such a model would be to assume that the occurence of a particular morpheme in a given context depends only upon the morpheme that precedes it. For the Turkish word `evimden`, the morphological decomposition is as follows:

```
(6)   ev    -im     -den
      house -POSS1s -ABL
      'from my house'
```

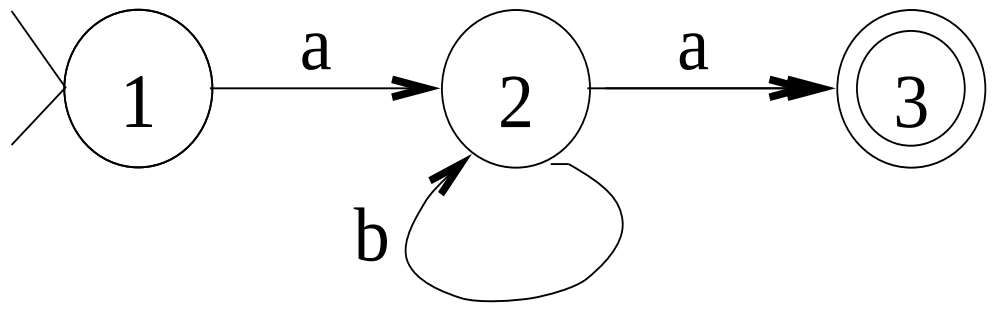

Figure III.2: An example of finite state automaton for $L_1$.

According to the nominal morphotactics of Turkish, the case markers (here the case is ablative) do not precede the possesive markers (here, the possesive marker is the first person singular). Thus, `evimden` is a valid word form, whereas, for instance, *`evdenim` is not a valid one. As a result, the morphotactics of languages which have concatenative morphologies can be adequately modelled by finite state automata. Figure III.3 shows a FSA that describes a fragment of Turkish nominal morphotactics.

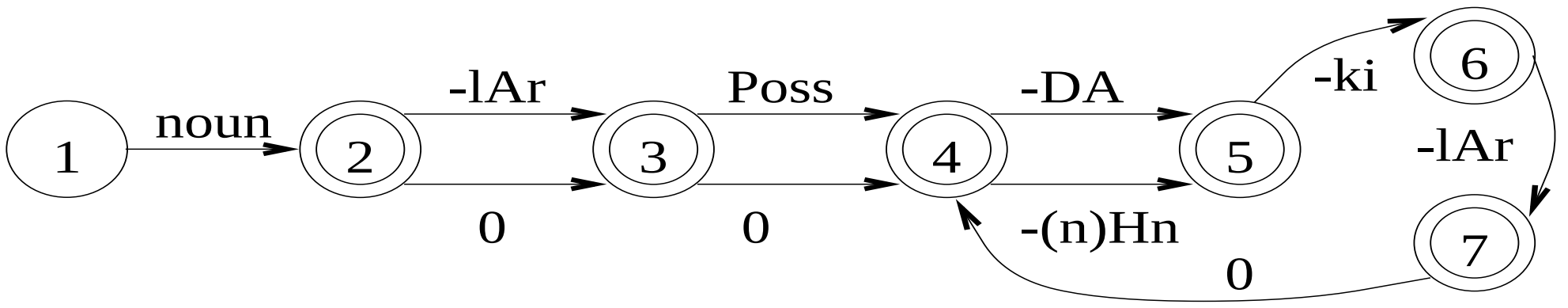

Figure III.3: An FSA for a fragment of Turkish nominal morphotactics.

Although the finite-state model is appropriate for handling the processes of concatenative morphology, it has some inadequacies. One such problem is the long-distance dependencies of concatenative morphology where the existence of one morpheme is allowed by another morpheme which is not adjacent to it. The following example from English will serve as an illustration.[1] The verbs formed by the prefix `en-` behave like normal English verbs in every respect, including the ability to take the suffix `-able`. e.g. `enjoyable`, `enrichable`. The suffix `-able` can attach when the prefix `en-` has already been attached. That is to say, the existence of `en-` before `rich` allows for the derivation of `enrichable`. In a finite-state model, one would have to have two different lexicon entries for each stem: one entry would represent the case where the prefix `en-` had not been used, and not allow `-able` to follow; the other entry would be for the case where the prefix `en-` had been used, and `able` would be allowed to follow. The same problem arises in the reduplication process of Turkish (see Subsection V.1.5). The problem discussed here turns out to be an inadequacy of some well-known finite-state models of morphology, in particular the original two-level model.

**Finite-state transducer**

A finite state transducer (FST) is like a FSA except that it simultaneously operates on two input strings. It recognizes whether the two strings are valid correspondences of each other. Thus a FST has an input and an output language. For instance, assume that the first input string is from language $L_2$ and the second input string is from language $L_3$. Below, two example correspondences which can be recognized by the FST in Figure III.4 are given.

[1] The example is from [33]

| | | | | |
|---|---|---|---|---|
| $L_2$: | aba | | $L_2$: | abcba |
| | | and | | |
| $L_3$: | aca | | $L_3$: | acaca |

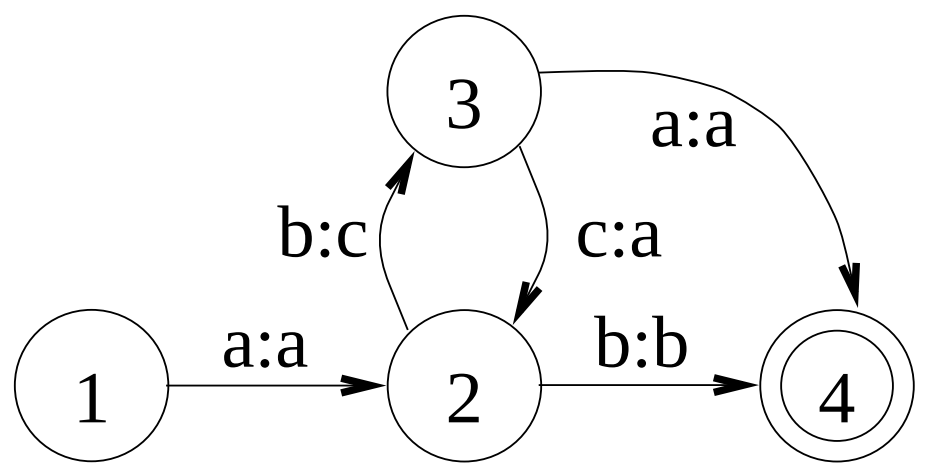

Figure III.4: FST diagram for the correspondence between languages L2 and L3

Note that, the only difference of a FST diagram from a FSA diagram is that the arcs are labeled with a correspondence pair consisting of a symbol from each of the languages. Whereas FSAs describe regular expressions, FSTs model regular relations [21].

A finite state model of phonology is especially desirable from the computational point of view, since it makes possible a simple and efficient computational implementation. Many linguists claimed that generative phonological rules could be implemented by finite state devices. Although the idea that the phonological rules can be modelled as FSTs has its origin in Johnson's work [15], the first proposal that the ordered rewriting rules of generative phonology could be implemented computationally as finite state transducers was made by Kaplan and Kay [17]. Each FST, in their aproach, would compare two successive levels of generative framework: the level immediately before application of the rule, and the level after the application of the rule. The whole phonological structure would then be a cascade of such levels and FSTs. The overall look of Kay and Kaplan's idea is sketched in Figure III.5.

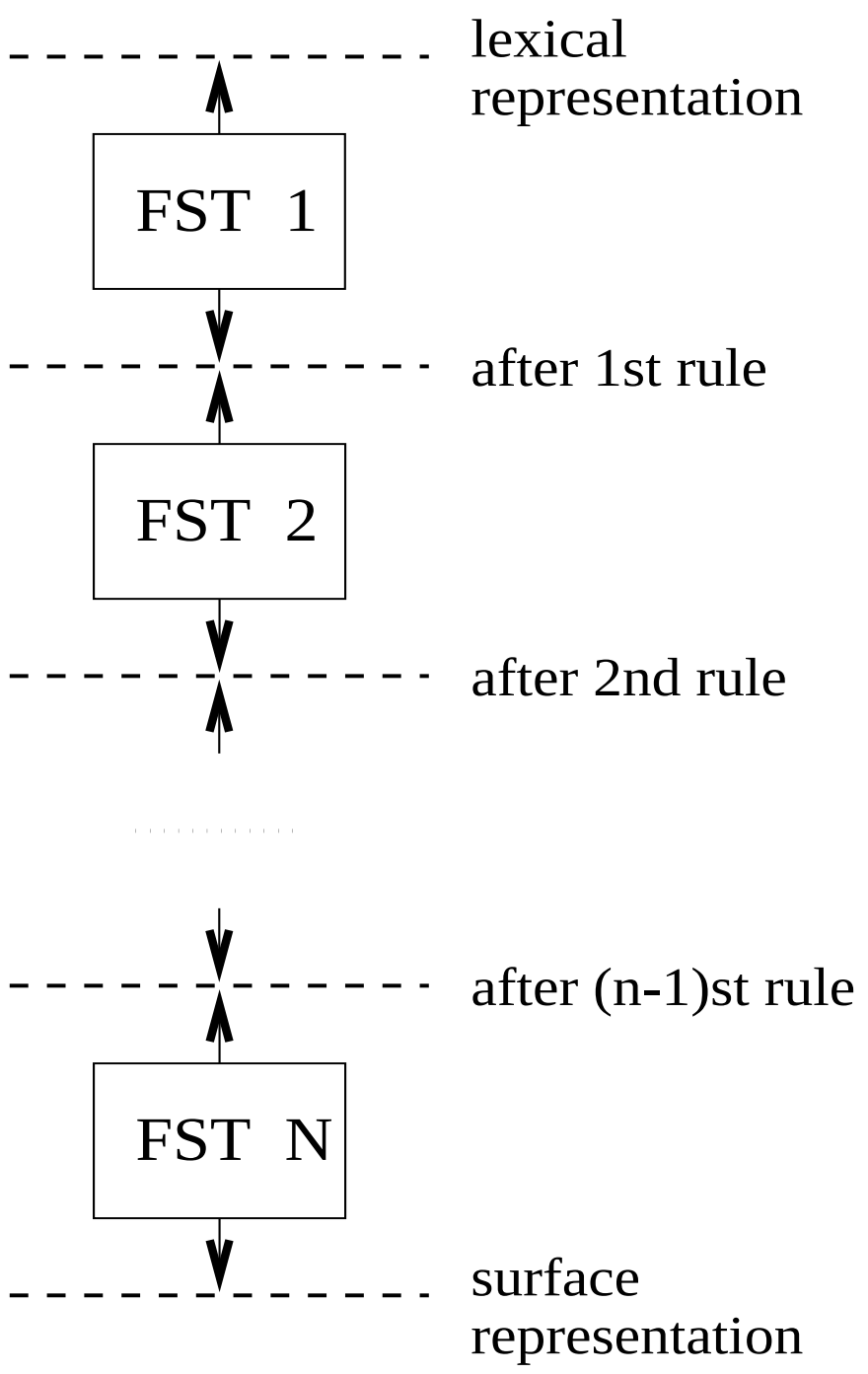

Figure III.5: Kay and Kaplan's idea: A cascading sequence of FSTs

However, a cascade of automata is not operational. But Kay and Kaplan suggested that the set of transducers could be merged into a single, larger transducer with the same input/output behaviour by using the techniques of automata theory. The merged transducer would be operational and bidirectional. The FST so composed, in the worst case, would have as many states as the product of the number of states of the individual machines. Although in most cases, the number of machines could actually be reduced by standard minimization techniques [12], the size of the merged transducer would be prohibitively large especially for the languages with complex morphology. Koskenniemi tried to apply this idea to the development of a working system, but having realized that it would be infeasible, he decided to model the phonological rules as a bank of automata running in parallel rather than in series. In this system, *every* FST sees both the lexical and the surface tapes, and there are only those two tapes. An overall look of two-level model is given in Figure III.6. Note that, when a machine is checking the correspondences between, say, lexical `y` and surface `0`, all other machines are also checking that correspondence.

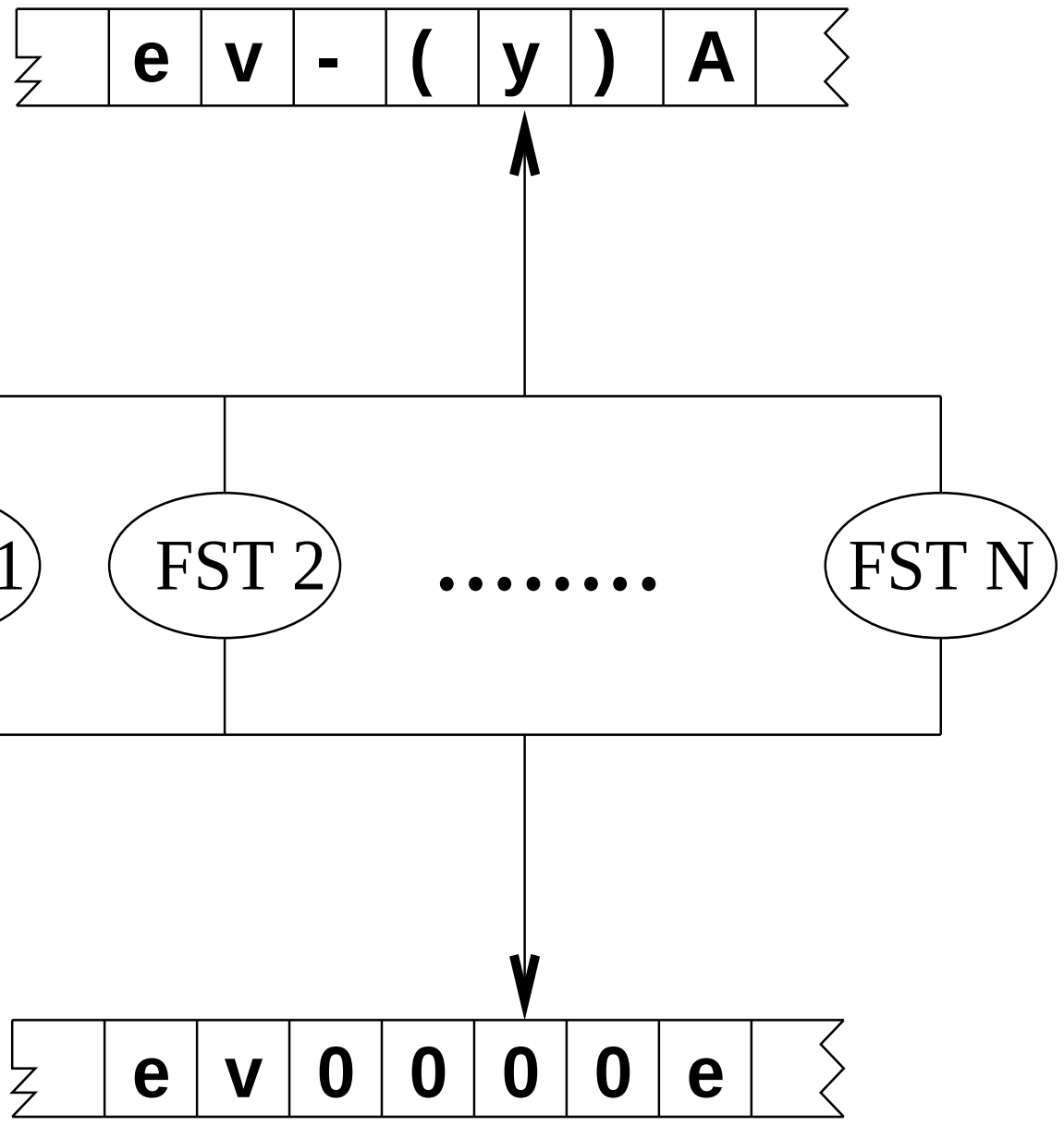

Figure III.6: The basic architecture of the two-level model: FSTs in parallel.

Figure III.7 shows how the process of a suffix-initial consonant drop can roughly be implemented using a FST. Here, only the deletion of the optional consonant, `(y)`, is presented. This segment constitutes the initial part for both the accusative, `-(y)H`, and the dative, `-(y)A`, suffixes of Turkish. Some examples of this process are:

```
(7) a. adam-(y)A →  adama
       man-DAT      'to man'

    b. kuş-(y)H →  kuşu
       bird-ACC    'bird (object)'
```

In the FST below, the states 1,2, and 6 are final. Some transitions from the intermediate states are ignored for the sake of simplicity. The @:@ arc[2] at the initial state allows any pairs to pass successfully through the FST except C:C (where C represents the set of consonants, and C:C, the default correspondences such as b:b, c:c, ç:ç, d:d, ...,z:z). The correspondence, `y`:0, maps the lexical symbol `y` to the surface symbol `0`[3]. That is, `y` is deleted. Here, the optional parentheses `(`, `)` and morpheme-boundary symbol `-` also corresponds to a surface null symbol.

[2] "@" is the ANY symbol.

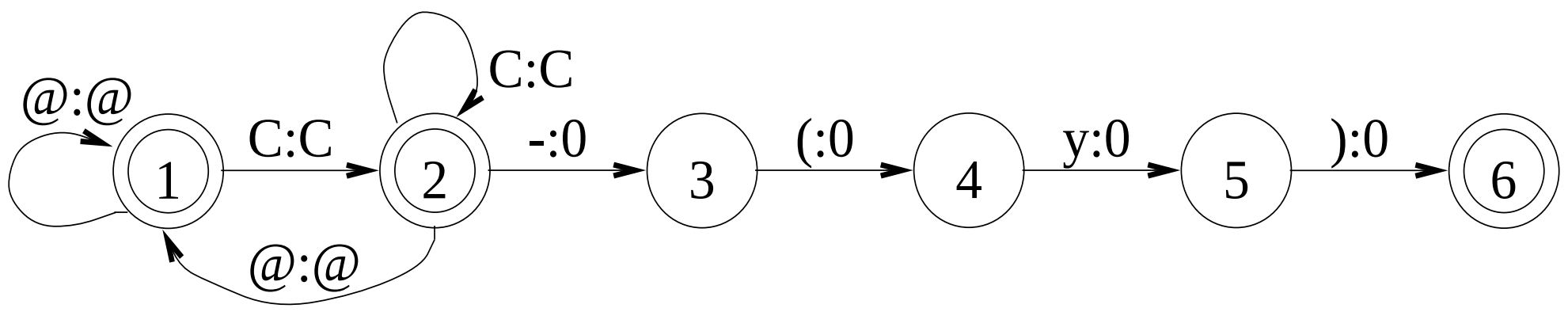

Figure III.7: A preliminary model of the deletion process in Turkish using a FST

## III.1 Two-level rules

Let us begin by examining the formal properties of generative rules. Briefly speaking, generative rules are sequentially ordered rewriting rules. Rewriting rules are rules that change or transform one symbol to another symbol. For example, a rewriting rule of the form $a \rightarrow b$ interprets the relationship between the symbols $a$ and $b$ as a dynamic change with which the symbol $a$ is rewritten (i.e., turned into) the symbol $b$. This means that after this operation takes place, the symbol $a$ no longer exists, in the sense that it is no longer available to other rules. In linguistic theory generative rules are known as process rules. Process rules attempt to characterize the relationship between levels of representation (such as the phonemic and phonetic levels) by specifying how to transform representations

[3] "0" is the NULL symbol. Phonological processes that delete or insert characters are expressed in the two-level model as correspondences with the NULL symbol.

from one level into representations on the other level.

Generative phonological rules apply sequentially. This means that each rule creates as its output a new intermediate level of representation. This intermediate level then serves as the input to the next rule. As a consequence, the underlying form becomes inaccessible to later rules.

Generative phonological rules are ordered; that is, the description specifies the sequence in which the rules must apply. Applying rules in any other order may result in incorrect output.

As an example of a set of generative rules, consider the following rules given in Table III.1 that roughly models some phonological processes in Turkish.

Table III.1: A sample group of generative rules

| | | | | | |
|---|---|---|---|---|---|
| 1. | H | → | ı | / Vbu C0 - (C) _ | |
| 2. | y | → | | / C1 - ( _ | 'Initial /y/ Elipsis' |
| 3. | [-,(,)] | → | | | 'Default correspondences' |
| 4. | ç | → | c | / _ V | 'Stop Voicing' |

Rule 1 above represents a part of vowel harmony: `H` (which stands for the set of high vowels) becomes `ı` only if the vowel in the preceeding morpheme is `Vbu`, where `Vbu` stands for the subset of back and unrounded vowels. Note that, among the symbols used, `C` stands for the set of consonants and `C0` stands for zero or more consonants, `-` is the morpheme boundary symbol, `(` and `)` are the optional parantheses, and `V` stands for the set of vowels. Rule 2 states that suffix-initial, optional `y` is to be deleted when the stem it appends, ends up with a consonant. Rule 3 rewrites the markers like morpheme boundary, optional parentheses as the null character (i.e., deletes the markers). Rule 4 states that `ç` becomes `c` in the environment preceeding `V`. A sample derivation of forms to which these rules apply looks like this (where LR stands for Lexical Representation and SR stands for Surface Representation):

```
(8)     LR:       ağaç-(y)H
        Rule 1:   ağaç-(y)ı
        Rule 2:   ağaç-()ı
        Rule 3:   ağaçı
        Rule 4:   ağacı
        SR:       ağacı
```

Notice that in addition to the underlying and surface levels, an intermediate level has been created as the result of sequentially applying rules 1, 2, 3, and 4. The application of rule 1, for instance, produces the intermediate form `ağaç-(y)ı`, which then serves as the input to rule 2. Note that, not only are these rules sequential, but they are ordered, such that rule 1 must apply before rule 2. Rule 1 feeds rule 2. Consider what would happen if they were applied in the reverse order. Given the the input form `ağaç-(y)H`, rule 1 will delete `y` producing the output form `ağaç-()H`. However, this time, rule 2 will not fire, since its environment is not satisfied.

### III.1.1 Two-level rules vs. generative rules

The example given in Table III.1 which demonstrates the usage of generative rules will frequently be refered to emphasize the differences between two-level rules and ordered rules. Two-level rules differ from generative rules in the following ways:

First, two-level rules apply in parallel. Applying rules in parallel to an input form means that for each segment in the form all of the rules must apply successfully. The two-level rules that produce the same surface representation as that of the generative rules used in the above example, may be as in Table III.2.

Table III.2: A group of two-level rules for a portion of Turkish phonology

| | | |
|---|---|---|
| 1. | H:ı ⇒ Vbu $C^*$ -:0 [(:0 y:@ ):0]$^*$ _ | 'A part of vowel harmony' |
| 2. | y:0 ⇔ C -:0 (:0 _ ):0 | 'Suffix-initial y elipsis' |
| 3. | ç:c ⇔ _ [-:0 (:0 y:@ ):0]$^*$ V | 'Stop voicing' |

There is no ordering relation among two-level rules. All of the rules operate simultaneously.

Second, whereas sequentially applied generative rules create intermediate levels of derivation, simultaneously applied two-level rules require only two levels of representation: the underlying (or lexical level) and the surface level. There are no intermediate levels of derivation. It is in this sense that the model is called two-level. For instance, in the previous subsection, a group of generative rules had rewritten the input form into a surface form after a number of intermediate steps. However, the two-level model would treat the same relationship between the lexical form and the surface form as a direct, symbol-to-symbol correspondence as follows:

```
(9)    LR: a ğ a ç - ( y ) H
       SR: a ğ a c 0 0 0 0 ı
```

Third, whereas generative rules relate the underlying and surface levels by rewriting lexical symbols as surface symbols, two-level rules express the relationship between the lexical and surface levels by positing direct, static correspondences between pairs of lexical and surface symbols. After a two-level rule applies, both the underlying and surface symbols still exist.

Finally, whereas generative rewriting rules are unidirectional (that is, they operate only in a lexical to surface direction), two-level rules are bidirectional. Two-level rules can operate either in a lexical to surface direction (generation mode) or in a surface to lexical direction (recognition mode). The practical application of bidirectional phonological rules is obvious: a computational implementation of bidirectional rules is not limited to generation mode to produce words; it can also be used toin recognize (parse) the words.

### III.1.2 Two-level rule notation

The two-level rules have the following basic syntax:

| | | | |
|---|---|---|---|
| *CP* | **op** | *LC* _ *RC* | `'a generic two-level rule'` |
| `y:0` | ⇔ | `C -:0 (:0 _ ):0` | `'a sample two-level rule'` |

*CP* stands for the correspondence part, and is a regular expression over the alphabet of feasible pairs, though it frequently consists of just a single pair. A feasible pair is a specific correspondence between a lexical alphabetic character and a surface alphabetic character. For instance, the first part of the sample two-level rule above is the correspondence `y`:0. The second part, **op**, is the rule operator. In the sample one, it is ⇔ . *LC* and *RC* stand for the left context and the right context respectively. These constitute the *environment* in which the correspondence occurs. These are also regular expressions over feasible pairs. The rule operator specifies the relationship between the correspondence and the environment in which it occurs. There are four types of operators and correspondingly four rules. These are:

`Context restriction rule:  a:b` ⇒ `LC _ RC`

`Surface coercion rule:  a:b` ⇐ `LC _ RC`

`Composition rule:  a:b` $\Leftrightarrow$ `LC _ RC`

`Exclusion rule:  a:b` $/\Leftarrow$ `LC _ RC`

The operator $\Rightarrow$ states that the correspondence `a:b` "only" occurs in the environment, but does not "always" occur. The operator $\Leftarrow$ states that `a` is "always" realized as `b` in the environment, but "not only" in that environment. The operator $\Leftrightarrow$ dictates that the correspondence `a:b` "always and only" occur in the given environment. This composition rule is actually a combination of a context restriction rule and surface coercion rule. Finally, the operator $/\Leftarrow$ states that, `a` is "never" realized as `b` in the environment.

Turning to the sample rule given above, notice that, there is a long underscore. It indicates the gap where the correspondence `y`:0 occurs. Such correspondences where the surface character differs from the lexical character are called special correspondences. These are mostly dictated as a consequence of a rule. Other correspondences such as `a`:`a` and `':'` are default correspondences. The sum of default and special correspondences makes up the set of feasible pairs. All feasible pairs of the description must be explicitly declared.

Correspondence part and environment part of a two-level rule consists of regular expressions over feasible pairs. Therefore, symbols for regular expression such as +, *, [, ], - etc. are used in writing two-level rules.

## III.2 Tools that use the two-level model

There is the issue of whether two-level description is a linguistic tool or a theory. A linguistic tool is used to describe natural languages. A linguistic theory, on the other hand, is intended to define the class of possible natural languages. From this point of view, two-level phonology is best regarded as a linguistic tool rather than a theory. Its job is to provide the expressive power needed to describe the phonological issues of natural languages.

Various computer implementations of the two-level model are available. The first implementation was in Pascal language and running on large systems. Then, Karttunen and his students implemented the model in INTERLISP and have made two-level descriptions of various languages. Initially two-level rules making up the phonological part of the model were coded by hand as FSTs. Although hand-coding the rules as FSTs is not very hard once one gets used to it, and is a good exercise for anyone interested in computational phonology, it gets very tedious if one wants to describe a language with a large number of rules. Thus, a compiler that could take two-level rules in the notation described above, has been often called for. Such a rule compiler was reported by Koskenniemi in 1986 and implemented (called *twol*) in the INTERLISP/D environment on Xerox workstations by Dalrymple et al. in 1987. Then, in 1990, *PC-KIMMO*, a computer program that runs the two-level model on both personal computers and other multi-user systems[1] was published by the Summer Institute of Linguistics (*SIL*). PC-KIMMO provides a general computational machine that executes a language description written by the user. A PC-KIMMO description of a language consists of two files. A rules file, which specifies the alphabet and phonological rules and a lexicon file, which lists lexical items (words and morphemes) and encodes morphotactic contraints. SIL also provides a rule compiler called *kgen*. There are also commercial products that use two-level model as a linguistic tool. One of them is the Two-Level Rule Compiler *twolc*[37] of Xerox, developed by Kart-tunen and Beesley. It is a compiler, written in C, that converts two-level rules to deterministic, minimized finite-state transducers. There is also a Finite-State Lexicon Compiler called *lexc*[25] designed by Karttunen at Xerox to be used in conjunction with transducers produced by the Xerox Two-level Rule Compiler. *Lexc* is an authoring tool for creating lexicons and lexical transducers.

## III.3 Discussion of two-level model

Two-level model has been by far the most successful model of computational morphology. The computational mechanisms that it uses for handling phonotactics and morphotactics are applicable to a wide range of languages. Another distinguished feature of two-level model is that it has a finite-state mechanism for phonology. There is no doubt that two-level model overpowers generative phonology in the sense that it can be computationally implemented. Yet, it has a limited level of observational and descriptive adequacy, and therefore, it is accepted as a tool more than being a theory.

Recall that two-level model requires a natural language description in two components completing each other: phonological and morphotactical component. Thus limitations can be examined from the points of these two components. The limitations concerning the phonological component will be expressed before these of morphological component.

In the initial implementations of the two-level model, the two-level rules were being handcoded. This was a severe inadequacy. Afterwards, rule compilers (See III.2) that effectively hide the inner workings of two-level model's finite-state machinery from user, have been developed. These compilers also use minimiza-tion algorithms adopted from the formal language theory and thus enhances the efficiency of the model.

Probably the most obvious sense in which the two-level framework is inadequate is that it is hard to build analyzers that handle nonconcatenative forms of morphology, such as infixation, gemination, and reduplication. Typical non-finite state solutions to problems like these involve variables that copy an element specified by a corresponding variable. Finite state machinery has no such mchanism for matching variables. However, the same effect can be achieved by enumarating all possible segments that result from the application of the process. Therefore, one can describe this kind of morphology, but only at some cost. For instance, Antworth [1](pp.151-162) discusses how to handle nonconcatenative processes using PC-KIMMO. However, Antworth notes that, tables that implement nonconcatenative morphology "often get very large and run slowly". One other type of nonconcatenative morphology is the templatic morphology found in Arabic. Many researchers claim that a fully developed description of Arabic within the two-level framework would be very complicated. There have been a few different proposals for handling the templatic morphology. The best-known attempt is that of Martin Kay [18]. His suggestion was to replace the single lexical tape of two-level model with three tapes.

As mentioned before, the strength of two-level model is in its phonological component. Concatenative morphophonemic processes can be implemented very effectively. However, the lexical component which stores lexical items and enforces morphotactic constraints is not as expressive as the phonological one. First, the lexical component encodes morphotactic information by specifying for each lexical item the class of items that can follow it. This simple approach crashes when there are cooccurence restrictions among morphemes that are not contiguous to each other, since whole sublexicons must be duplicated in order to represent different morphotactic sequences. The source of this problem lies in the fact that tools within the two-level framework encodes morphotactic information directly in the lexical entries themselves. A better approach would be to state morphotactic constraints in rules that are separate from the listing of lexical entries. For example, consider how a description of English should allow the words `enjoyable` and `enjoy` but disable the nonword `joyable`. Apart from duplicating the word `joy` with two different continuation, one could partially solve this problem by using some markers that would designate the occurence of invalid strings and block the corresponding two-level rule.

One solution, as Sproat[33] reports from Bear [5], is to implement a *unification scheme* such as the one in PATR formalism, in other words, a word grammar which will handle morphotactics. Beesley[6], having adopted Bear's idea, uses a unification scheme to handle Arabic discontinuous verbal morphology. Recently, PC-KIMMO version 2 has been published by SIL with new features. Among these new features, the word grammar component is the most significant. The word grammar component, like Beesley's work, uses a unification-based parser based on the PATR-II [31, 32] formalism.

Another problem with a word such as `enjoyable` is that the lexical components of the tools cannot determine the lexical class (part of speech) of the whole word. This is because the gloss of a recognized form is merely a concatenated string of the glosses of the morphemes that comprise the form. Because the gloss is a flat sequence of labels, it does not express the internal bracketed structure of the word. This deficiency made the tools using two-level model less usable as a front-end to a syntatic parser, since a syntactic parser must know the category of each word. Therefore, the output from a morphological analysis of a word form has to be postprocessed. It is claimed that the recently published PC-KIMMO version 2 also provides a full feature specification for a word form with the feature passing mechanism through the use of the word grammar component. It will not be unfair to state here that these limitations of the two-level model have been faced by the author during the design and implementation phases of this study. The personal comments about these limitations will be discussed in the following chapters.

In general, the two-level model is appropriate for languages with concatenative morphologies. It is strong in phonology and relatively weak in morphology.

### III.3.1 Complexity of the two-level model

Although Koskenniemi originally reported that the two-level processor runs in linear time due to the fact that finite-state machines are very fast, the two-level model uses an exponential algorithm as Barton et al. [3, 4] showed. At any point in parsing (or a generation) there are as many paths to explore as there are alternations involving the current surface (or lexical) symbol. Thus finite-state machinery does not guarantee efficient processing, and two-level parse (or generation) is NP-hard. Barton et al. [4] (pp. 155-159) prove the NP-hardness of two-level generation showing that one can encode the *Boolean satisfiability* problem as a two-level generation problem. Hence the two-level generation is generally intractable. He and others also designate the complexity issues of a standard two-level implementation and propose solutions to the problems. One of these solutions involves merging the many lexicons into one so that only one lexicon must be searched.

Koskenniemi and Church [21] responded to the claim of Barton et al. by stating that the potential for exponentiality is never likely to be realized in any practical application of the model for real data, because there is no natural language that posses complexity features compatible with satisfiability problem. Hence, practical implementations of two-level model will work in linear time.

## III.4 Other tools for computational morphology

Two-level model has an important place in computational morphology, yet, there are some systems such as DECOMP [26] which predates two-level model by many years.

MIT's DECOMP module, being one of the earliest morphological analyzers, was developed in connection with the work on the English text-to-speech system MITalk. It had a finite-state model of morphotactics and a very simple model of spelling changes.

The next finite-state morphological analyzer that is worth to be mentioned is Hankamer's `Keçi` system [11] for Turkish. Hankamer's system has a strictly finite-state morphotactics such as two-level model and DECOMP. What is particularly interesting about `Keçi` is the treatment of phonological rules. It uses a generate-and-test procedure incorporating ordered phonological rules in the style of generative phonology. It is reported by Sproat [33] (pp.191-192) that due to the weaknesses of generative phonology as opposed to that of two-level phonology, `Keçi` is more limited than two-level systems in what it can accomplish.

There are also works where the morphotactics is explicitly non-finite-state (i.e., context-free). One of them is AMPLE [1], an earlier contribution of SIL. AMPLE models morphotactics with a version of categorial morphology.[4] AMPLE has mechanisms that allow long-distance dependencies between affixes and thus, in principle, have greater than finite-state power AMPLE has no direct model of phonological rules, and it is therefore necessary to list all the surface forms in which a morpheme might occur.

General trends in the computational morphology literature can be summarized as follows:

- Most morphological analyzers use finite-state morphotactics inspite of the existence of some context-free systems.

- In the majority of systems, all morphemes are considered to be dictionary entries. That is they use item-and-arrangement model.[5]

[4] This categoral morphology is same as the kind of morphology presented in the word grammar component of PC-KIMMO version 2.

[5] where a word is assumed to be built up by the successive addition of morphemes to stems of increasing complexity, all morphemes are lexical items meaning that just as the root `çay` ("tea") is put in the lexicon, the suffix `-CH` (derives the word `çaycı` "one who prepares and sells tea" when attached to the root) has to be put in the lexicon.

# CHAPTER IV

## TURKISH

Turkish is a member of closely related *Turkic* languages and belongs to the southwestern group of the Turkic family with its sister languages `Azeri`, `Gagavuz` and `Afşar` Turkish. Turkic languages are in the *Altaic* branch of Uralic-Altaic Language family.

In terms of word formation, Turkish is considered an agglutinative language; morphemes are clearly identifiable in the make-up of the word. Phonological alternations in morphemes are also reflected in orthography in many cases. The following is an example of voicing (`ç` → `c`) (in Turkish: The word `ağaç` "tree" turns into `ağaca` "to the tree" when it takes the inflectional suffix `-(y)A`. In Turkish, as in Finnish, the ortography is indeed almost a perfect phonemic transcription. As a contrary example (at least for languages which use alphabetic writing systems) to Turkish, English is at the other extreme, where there is poor correspondence between pronounciation and orthography. Therefore, morphological analyzers for English typically deal with orthographic rules which often do not correspond to phonological rules of English.

## IV.1 Morphology

Turkish is famous for her rich phonological features. These features impose some phonological processes such as assimilation, reduction, and etc. in the contexts satisfying certain conditions. Vowel and consonant assimilations are the most typical features of Turkish. Almost all the phonological processes and alternations in Turkish occur as a consequence of the *minimum effort* principle. These processes can generally be formulated as (morpho)phonological rules (i.e., *morphophonemics*). The term generally is used because there are quite many exceptions violating the phonological features of Turkish. Almost all of these exceptions are loaned words. The others are native ones which had experienced phonetic changes because of various reasons. Yet, the productivity of Turkish has been preserved by some rules of exception or explicitly, by listing the exceptional words. The phonological processes of Turkish will be briefly mentioned in detail in the following sections. Phonetic features of Turkish will not be covered unless the phonetic information affects the morphophological process since this work does not deal with the sound synthesis or analysis.

### IV.1.1 Turkish phonemes

There are 8 vowels `a`, `e`, `ı`, `i`, `u`, `ü`, `o`, `ö` and 21 consonants `b`, `c`, `ç`, `d`, `f`, `g`, `ğ`, `h`, `k`, `j`, `l`, `m`, `n`, `p`, `r`, `s`, `ş`, `t`, `v`, `y`, `z`. Tables IV.1 and IV.2 show the features of vowels and consonants respectively.

It has been stated that in Turkish, the orthography is almost a perfect phone-

Table IV.1: The layout of Turkish vowels

| | a | o | ı | u | e | ö | i | ü |
|---|---|---|---|---|---|---|---|---|
| high | - | - | + | + | - | - | + | + |
| round | - | + | - | + | - | + | - | + |
| back | + | + | + | + | - | - | - | - |

Table IV.2: The layout of Turkish consonants

| | | Bilabial | Labio-dental | Dental Alveolar | Plato Alveolar | Palatal | Velar | Glottal |
|---|---|---|---|---|---|---|---|---|
| Stop | Voiceless | p | | t | ç | | k | |
| | Voiced | b | | d | c | | g | |
| Fri. | Voiceless | | r | s | ş | | | |
| | Voiced | | v | z | j | | | |
| Nas. | | m | | n | | | | |
| Liq. | Lateral | | | l | | | | |
| | Nonlater. | | | r | | | | |
| Glide | | | | | | y | ğ | h |

mic transcription. However, there are inevitable differences between what is written and pronounced. That is to say, there are some sound variants (i.e. allophones) of a phoneme that are represented by the same symbol of the phoneme. The sound variation is established during pronounciation by the effect of neighbouring sounds. For instance, the letters `k`,`g`,`ğ`, and `l` correspond to two phonemic variants according to the place of articulation. Among these `k`,`g`, and `ğ` are velar and are used with the back (`kalın`) vowels.[1] Their palatal) cognates are used with the front (`ince`) vowels.[2] The same backness feature[3] of the tongue applies also for the phoneme /l/. It has two allophones: Dental velar when the tongue is at the back[4] and dental alveolar for the other case. Naturally, the dental alveolar

[1] e.g., `kadın`, `dağ`, `gaga`.

[2] e.g., `keser`, `yelek`, `gemi`, `lig`, `iğne`.

[3] For a detailed explanation of the SPE system of distinctive features, see [16].

[4] Note that because in SPE the position of tongue in the production of a mid front vowel is taken as the neutral position, it is not only back vowels that are [+*back*], but also central ones due to the binary feature of the system.

/l/ is used with front vowels (e.g., `kil` "clay") and the dental velar one with back vowels (e.g., `çatal` "fork"). There are also some long vowels mainly used in words borrowed from foreign languages, most notably Arabic and Persian. Such vowels are sometimes distinguished in older orthography by various means (such as with a ˆ on top of the vowel). Although no extra orthographic symbol is used to represent these sound variants in modern orthography, some will be introduced to designate the proper environments while formulating the phonological processes.

### IV.1.2 Vowels and consonants

The number of vowels in Turkish is more than that of many languages and almost every morpheme[5] contains a vowel.

Vowels of Turkish can be classified in three groups according to their articulatory properties. These properties had been mentioned in Subsection II.1. A first distinction is between *front* and *back*: front vowels belong to the set (`e, i, ö, ü`) and back vowels (`a, ı, o, u`). The second one is being *round* and *unrounded*: round vowels are (`o, ö, u, ü`) and unrounded vowels are (`a, e, ı, i`). The third factor determines the characteristic of being *high* (close) or *low* (open), in other words the sets (`ı, i, u, ü`) and (`a, e, o, ö`) respectively. Vowel harmony which is a kind of sound assimilation is formulated using these vowel subsets.

In Turkish, the low rounded vowels `o` and `ö` can only be used in the first syllable of a word but not in the other syllables; and there is no suffix that contains a low rounded vowel apart from the verbal suffix `-(H)yor`. In some compound (e.g., çöreotu) and borrowed (e.g., doktor) words, o and ö can be seen in the other syllables. Note that, the word roots containing two adjacent vowels are borrowed words (e.g., şiir "poem", saadet "happiness").

[5] Some bound forms do not have a vowel: a variant of causative suffix `-t` and a group of personal suffixes `-m`, `-n`, `-k`.

There are no diphthongs in Turkish, but there are some other vowels apart from the ones in the alphabet. Some of these are only used in spoken language and others exist in the loan words and even used in the written language. Here, the ones which are effective in phonological processes and/or used in written language will be listed:

- There is a primordial high vowel, called *high* e. It is actually a sound between e and i and will be denoted as è. It had been in use during the old and middle phases of Turkish[6] and still exists in the first syllable of some words in some Anatolian dialects. Some examples of these words are yèmek, dèmek, yètmek, èl, yèr, gèce, èrken, and vèrmek. Today, it is pronounciated as the low e (in İstanbul dialect which is accepted as the standard) and written as such (i.e., erken, yetmek, etc.). However, the existence of è will be preserved in the words significant for the formulation of exceptional cases. The first two of these are discussed next.

  The words yèmek and dèmek are both in the form Root+Infinitive(-mAk). The verb roots are yè and dè respectively. When these verbs take suffixes of future and continuous aspect, è in the first syllable maps to i. On the other hand, when these words are inflected by other tense/aspect suffixes, è becomes e. For instance,

[6] See [7] for the evolution of Turkish

| | | |
|---|---|---|
| `yè`+Future(`-(y)AcAk`) | → | `y`*ı*`yecek` |
| `dè`+Continuous(`-(H)yor`) | → | `d`*ı*`yor` |
| | vs. | |
| `yè`+Past(`-dH`) | → | `y`*e*`di` |
| `dè`+Optative(`-mAlI`) | → | `d`*e*`meli` |

- The traces of an acute (front) `a`, which will be denoted as `á` can still be observed in some of the loan-words from Arabic such as `saát`, `hárb`, `hárf` etc. These are written by using the standard back `a`, but pronounced like a front `a` (`á`). This frontness feature shows its effects in vowel harmony. The acute `á` phoneme in the last syllable of the morpheme of the loan-base assimilate the following vowel of the suffix in respect to the frontness feature. Written form seems like a contradiction in the surface level. For instance,

| | | |
|---|---|---|
| `harb`+LOCATIVE(`-DA`) | → | `harbte` |
| `harf`+ACCUSATIVE(`-(y)H`) | → | `harfi` |
| `saat`+DATIVE(`-(y)A`) | → | `saate` |

Consequently, `á` will be used in loan-words (i.e., `saát` instead of `saat`) so as to mark the context of exception.

There are long vowels in Turkish, or if there had been any, they are extinct by now. Yet, the feature of length may be required for contrasts in inherited Arabic and Persian long vowels. The vowels in the first syllables of words like `alim`, `munis`, `şive`, etc. and in the final syllables of words like `rica`, `arzu`, `vefa` etc. are pronounciated as long vowels but written in accordance with Turkish orthogra-phy. The sequence Vowel+ğ+Consonant also yield Long-Vowel+Consonant like in /ağzı/ → [a:zı]. However, length is phonetic and therefore out of the scope of this work.

The features of Turkish consonants are given in Table IV.2. The feature of being voiced "yumuşak" or voiceless "sert" should be marked since it is effective in consonant harmony. Voiceless consonants are ç, f, h, k, p, s, ş, and t; and the others are voiced. The consonants f and j are not originally found in Turkish. Therefore, words containing and especially beginning with these consonants are few. Words such as spor "sport", tren "train", klasik "classical" and etc. that begin with two consonant clusters do violate the syllabification paradigm of Turkish and are not originally Turkish. A syllable may end in two main series of consonant clusters. The first type of such a syllable-final consonant cluster consists of a sonorant plus and obstruent (e.g., lç, lk, lp, lt, nk, nt, rç ,rk, rp, rt) as in the words ölç "to measure", renk "color", genç "young", sert "tough", and etc. The second type of syllable final cluster is made up by a voiceless fricative plus an oral stop (e.g., ft, st) as in words çift "pair", dost "friend".

## IV.2 Phonological and morphophonemic processes

As mentioned before, Turkish is a language of harmony. Therefore, anyone who will make a morphological tool for Turkish must model the phonological processes. These phonological processes should be seen from two points of view: *Phonemics*, the general rules of the grammar which are valid for all parts of the string fulfilling certain phonological requirements regardless of the identity of the individual morphemes, and, *morphophonemics*, the assignment of correct *allomorphs* for certain suffixal and some base morphemes. Sound alternations in Turkish arise out of various reasons and can be expressed in terms of the realization of minimum effort principle. Some of the major reasons are: the existence of some exceptional words (usually the borrowed words of the lexicon) that do not fit in the norms of Turkish phonology, invalid ordering of vowels in stems or within morpheme junctures, and the assimilation of two neighbouring sounds that have distinct phonetic features such as the place of articulation.

These alternations are generally phonemic and take place in the root or stem. Some of these alternations are effective in the written language. Others are observed only in the spoken language, specifically in some local dialects. Some examples of vowel and consonant assimilations that occur during pronounciation are given below:

(10) a. ş*o*för → ş*ö*för 'driver'

b. yan*l*ış → yan*n*ış 'wrong'

c. an*l*amak → an*n*amak 'to understand'

d. e*c*zane → e*z*zane 'pharmacy'

e. gitme*z*se → gitme*s*se 'if he does not go'

### IV.2.1 Vowel harmony

Vowel harmony is the most interesting and distinctive feature of Turkish. It is a sound assimilation, and, like most of the phonological processes, a natural consequence of the minimum effort principle. Generally speaking, the vowel harmony is the assimilation of the vowel of a morpheme by the vowel of the preceeding morpheme based on the features of Flatness and Gravity (i.e. The first vowel assimilates the following vowel). Vowel harmony is *progressive.* As stated before, the articulatory properties of phonemes are effective in assimilation. The articulatory property that is in charge here is the place of articulation. Vowel harmony as a whole actually consists of two such assimilations: a) Palatal assimilation and b) labial assimilation.

a) Palatal assimilation (`büyük ünlü uyumu`)

   This assimilation depends on the feature of Gravity (*front* vs *back*). The description of the so-called "major vowel harmony" is as follows:

   i) If the vowel of the first morpheme in a word is back then the vowels of the following morphemes are back (e.g., `odun` "wood", `sıcak` "hot", `onarmak` "to mend").

   ii) If the vowel of the first morpheme in a word is front then the vowels of following morphemes are front (e.g., `ipek` "silk", `ekşi` "sour", `dilemek` "to wish").

   The "major vowel harmony" is common to almost all Turkic languages and languages of Uralic family.

b) Labial assimilation (`küçük ünlü uyumu`)

   The so-called "minor vowel harmony" assimilation depends on the feature of Flatness, i.e., the interplay between *rounded* and *unrounded.* This assimilation can be described as follows:

   i) If the vowel in the first syllable of a word is *unrounded* then the vowels in the succeeding syllables of the words are *unrounded* (e.g., `erken`

"early", `bilek` "wrist").

ii) If the vowel in the first syllable of a word is *rounded* then there are two possibilities:

- The vowels of the succeeding syllables should be *rounded* if they are *close* (high) (e.g., `ödül` "prize", `kutu` "box").
- The vowel of the succeeding syllable remains *unrounded* if it is *open* (low) (e.g., `güneş` "sun", `yumurta` "egg"). The second case is a contradiction for labial assimilation. If it were not, the above words should be realized as `günöş` and `yumurtu`.

Although the vowel harmony is a phonemic process, it is modelled morphophonemically due to some practical reasons: First, the words of the borrowed lexicon usually violate the vowel harmony constituting a group of irregular roots. Second, almost all the suffixes harmonize with the roots whether they are native or borrowed. Thus, the harmony rules are applied to the suffixes at the morpheme (root, word, suffix) boundaries yielding to a group of allomorphs for each distinct morpheme type. This hormonization process is progressive which leaps over morpheme boundaries and enforces assimilation in the successive morphemes.

### IV.2.2 Consonant harmony

Consonants can be classified according to some distinctive feature sets. The most effective distinction is between being voiced or voiceless, and consonants in Turkish mostly harmonize according to this distinction. As a principle, consonant clusters, even at morpheme boundary, are either totally voiced or totally voiceless.

Note that harmonization process is effective for all parts of strings constituting root, stem, and morpheme bodies. Therefore, consonant harmony is actually a phonemic process rather than a morphophonemic process. Borrowed words that violate consonant harmony in their original forms agree with the harmonization process during the realization of their Turkish counterparts and become lexicalized. Some examples for such borrowed words are listed below.

| | | | | | |
|---|---|---|---|---|---|
| (11) | a. | ıztırab | → | ıstırap | 'suffering' |
| | b. | tesbit | → | tespit | 'assessment ' |
| | c. | mikdar | → | miktar | 'amount' |
| | d. | mektub | → | mektup | 'letter' |
| | e. | reng | → | renk | 'color' |
| | f. | Ahmed | → | Ahmet | |

Consonant harmony in Turkish cannot be accounted for by a single rule. A great number of borrowed and native words display an irregular character, which makes it hard to formulate the consonant harmony. At this point, some processes related with the consonant harmony will only be exemplified. Details of these processes, exceptional cases, and design issues will be presented in the next chapter (see V.1.2).

(12) a. reng → renk
color-NOM
'color'

b. ilac-tan → ilaçtan
color-ABL
'from the medicine'

c. ilac-ı → ilacı
medicine-ACC
'medicine (object)'

(13) a. ağaç-ı → ağacı
tree-ACC
'tree (object)'

b. sokak-a → sokağa
street-DAT
'to street'

c. gel-ecek-i → geleceği
come-DER-POSS3s
'his/her future'

Apart from the consonant assimilation that relies on the distinction between voiced and voiceless, there are also consonant assimilations lacking generality. Among these, the alternation n-m that occurs before bilabial stops (i.e., b, p) is the most frequently observed. Some of these, especially the ones that occur in the borrowed roots, are reflected in the written language.

(14) a. te*m*bih (< Ar. te*n*bih) 'order'
b. ca*m*baz (< Frs. ca*n*baz) 'acrobat'

Alternations like in the examples above had already become permanent and these words have been lexicalized. Such realizations are also out of the scope of this study. They are mentioned to give an insight about how the general phonological features of Turkish enforce restrictions over and cause changes in the borrowed words.

Apart from the harmony rules above there are other morphophonemic processes such as vowel and consonant epenthesis, reduplication. These will be expressed in detail in the next chapter. Besides, a number of minor morphophonological regularities in the borrowed vocabulary of Turkish allow some generalizations to be made. These processes of so-called *borrowed-word phonology* can also be generalized. Design and implementation issues for the formulation of these processes will be given in the next chapter.

## IV.3 Morphotactics

From a morphological point of view, Turkish is a perfect agglutinative language. Both inflectional and derivational systems apply the principle of stacking suffix to suffix. This may result in long words which may be the equivalent of a whole phrase, clause, or even a sentence in non-agglutinative languages.

```
(15) a. yurtdışı -nda -ki  -ler -den -miş -ler
        abroad   -LOC -REL -PLU -ABL -AUX PERS3p
        'They seem to be of those who are abroad'

     b. sor -uş   -tur  -ul   -muş   -um     -dur
        ask -RECP -CAUS -PASS -TENSE -PERS1s -COP
        'It is for sure that i had been investigated'
```

Turkish has two main paradigms for word formation. The *nominal paradigm* applies to nouns and adjectives and describes the order of the inflectional suffixes. Note that the adjective and noun distinction in Turkish is difficult to make and these two are sometimes collectively called *substantives.* Most adjectives can be used as nouns, and undergo the derivations from a noun. Nouns can perform the function of an adjective as noun modifier in noun-noun groups [36]. The *verbal paradigm* applies to verbs and describes the order of inflectional suffixes that are applicable to verbal stems. There are also cross-paradigm derivations that derive verbs from nouns and vice versa. These paradigms will be discussed thoroughly in the next chapter.

Turkish morphotactics allow productive formation of words whose part-of-speech categories may change a number of times during affixation. One can start with a nominal root, then form a verbal form with a suffix which then take an aspect suffix and then become a nominal form again through, for example, a gerund suffix, and then take the standard nominal inflectional suffixes (plural, possesive, case, etc.). It is also possible to have circular constructions; one example of this is the relativization suffix `-ki`. Apart from the constraints imposed by morphotactics on affixation, there is also semantic restrictions. The act of handling the semantic restrictions is a design and implementation issue. Actually, most of the morphological analyzers do not provide sufficient mechanisms to model such restrictions. In the following chapter, the design issues for morphophonemics and morphotactics will be covered.

# CHAPTER V

## DESIGN

Our lexical representation for affixes make two notions explicit. First, optional segments are designated by enclosing parentheses, e.g., `-(y)A` for dative suffix. Second, meta-phonemes (or archiphonemes) are represented by upper-case symbols, e.g., `-(H)mHz` for the first person plural possesive suffix, where `H` stands for `ı/i/u/ü` (high vowels). Optional segments and meta-phonemes are mapped to zero forms (phonologically null) or surface forms by the rules.

### V.1 Morphophonemic processes

During the design and implementation phase of morphophonemic processes, we concentrated on two principles.

1. Strive for a general purpose model and embed the maximimum amount of linguistic information into the rules.

2. Try to be as practical as possible but not violate the linguistic relevance of rules.

Note that the rules of vowel harmony, final-stop devoicing, suffix-initial devoicing/voicing, suffix-initial vowel and consonant drops are generally recognized as being productive, whereas the rules describing final-stop voicing, gemination, vowel deletion (insertion), reduplication must be considered as nonproductive. However productivity is not an absolute measure for the applicability of a rule in an implementation. A productive rule may stand for a general tendency but does not necessarily correctly predict form alternations in individual cases. These processes should be treated carefully.

### V.1.1 Vowel harmony

Although vowel harmony is a phonemic process and therefore, should be valid for all parts of strings constituting Turkish word forms, it is unpractical to formulate within the root since there are many disharmonic native and borrowed roots. Most of these are borrowed roots such as `elektronik`, `hilal`, `rüzgar`. Yet, there are also a few number of native ones such as `şişman`, `elma`, `kardeş`, `inanmak` which violate the major vowel harmony and such as `çabuk`, `kabuk`, `tapu`, `yağmur`, `avlu`, `avutmak` which violate the minor vowel harmony. Almost all these exceptions have taken their current forms due to various explainable reasons (e.g., *palatalization*: `alma` > `elma` and *labialisation*: `kabık` > `kabuk` due to the rounding effect of syllable initial labial consonants `b` and `p`). Therefore, the vowel harmony should be modelled as morphophonemic process in which the vowel in the last syllable of the root (whether an exception or not) assimilates the vowel of the suffix that is affixed. In other words, the vowel of the suffix agree with the vowel in the last syllable of the preceeding morpheme.

For the morphophonemic alternations, we use the meta-phonemes `A` and `H`. Thus suffixes can be divided in two groups: those which contain a *high/low* vowel `H` and those which are based on a *back/front* vowel `A`. Note that `A` corresponds to the restricted set (a,e)[1] whereas `H` maps to the full set (ı,i,u,ü).

Vowel harmony can be formulated using 6 distinct but similiar two-level rules for each of the realizations shown in Table V.1. Note that each set of vowels in the first column represents a dependency for each correspondence given on the right.

Table V.1: Correspondences for vowel harmony

| Preceding vowel | LR:SR |
|---|---|
| Vfr = {e,i,ö,ü} | A:e |
| Vbk = {a,ı,o,u} | A:a |
| Vfrurd = {e,i} | H:i |
| Vbkurd = {a,ı} | H:ı |
| Vfrrd = {ö,ü} | H:ü |
| Vbkrd = {o,u} | H:u |

Two-level rules for vowel harmony are given in Table V.2. There are some other rules that applies vowel harmony for some other meta-phonemes. `LCVH` denotes the left context for vowel harmony. It is a complex regular expression that defines the context between the correspondence and the preceding vowel. Note that this context should not block the operation of other rules. From the point of vowel harmony rules, `LCVH` actually consists of many "don't care" realizations. The operator $\Rightarrow$ , for instance in the first rule, dictates that `A` is realized as `a` only in this context but there may be another realization of `A` in the very same context. The second realization involves the case where `A` drops. Therefore, the operator ⇒ is used instead of ⇔ . Refer to the rules file in the appendix for a better understanding of the contexts and additional rules for vowel harmony.

[1] This is due to the fact that the low rounded vowels o and ö can only be used in the first syllable of native Turkish word and in no suffix except the verbal suffix `-(H)yor`.

Table V.2: Two-level rules for Vowel Harmony

| | | | | | |
|---|---|---|---|---|---|
| 1. | A:a | ⇒ | [:Vbk - Vac:] | SLCVH | _ |
| 2. | A:e | ⇒ | [:Vfr \| Vac:] | SLCVH | _ |
| 3. | H:ı | ⇒ | [:Vbkurd - á:a] | SLCVH | _ |
| 4. | H:i | ⇒ | [:Vfrurd \| á:a] | SLCVH | _ |
| 5. | H:u | ⇒ | [:Vbkrd - Vacrd:] | SLCVH | _ |
| 6. | H:ü | ⇒ | [:Vfrrd \| Vacrd:] | SLCVH | _ |

The first vowel harmony rule dictates that if the surface realization of the preceding vowel is back, then `A` corresponds to `a` otherwise to `e`. Note that `Vac` denotes the set of some acute variants.[2] Among these `á` was introduced in Subsection IV.1.1. Recall that `á` occurs in the final syllable of some borrowed-words such as **`dikkát`**, **`saát`**, **`hárf`**, and **`hárb`** and causes an acute assimilation, although its surface representation is `a` in accordance with Turkish orthography.

Here, let us introduce one regularity in harmonic assimilation which applies to non-native bases, namely: when a borrowed morpheme ends in a dental alveolar /l/, it imposes acute harmony on the following vowel. Banguoğlu ([2]:91-92) and Lees ([24]:52-53) discuss this harmony in detail. It also affects the preceding vowel forming a word-final acute context. Some examples of this acute assimilation are listed below.

```
(16) a. alkol   -lH  →  alkollü
        alcohol -DER
        'containing alcohol'

     b. kabul      -lAn →  kabullen
        acceptance -DER
        'accept'
```

[2] The set for the acute variants of back vowels is {`á, ó, ú`}

This is a memorized regularity for most of the native speakers. Therefore, in practice it is not convenient to use two different symbols to represent the allophones of /l/. As vowel harmony works upon the preceding vowel and the dental alveolar /l/ palatalizes the preceding vowel within this acute context, it is more practical to mark the vowel but /l/. The usage of the acute variants ó and ú, is therefore not erroneous to designate the existence of such an acute context in disharmonic roots, even though ó and ú are not a part of Turkish orthography. Such a representation has its basis in the linguistic tradition. Consequently, such borrowed disharmonic roots will contain above acute vowels in their lexical entries. Some of these are `dikkát`, `saát`, `usúl`, `futból`, `gól`, and `súlh`.

Oflazer [27, 28] in his work directly uses some markers (e.g., `V` for `á`, `^` for `ó` and `&` for `ú`) to formulate this process neglecting the underlying theoretical relation.

Note that meta-phonemes in the suffixes are resolved in progressive order. Two-level rules for vowel harmony operate by checking the validity of a correspondence by refering to the surface realization of the preceding vowel. Figure V.1 illustrates the progressive character of vowel harmony for the example (22).

```
(17)   ev    -(s)HN  -DA  -sHnHz
       house -POSS3s -LOC PERS2p
       'you are in his house'
```

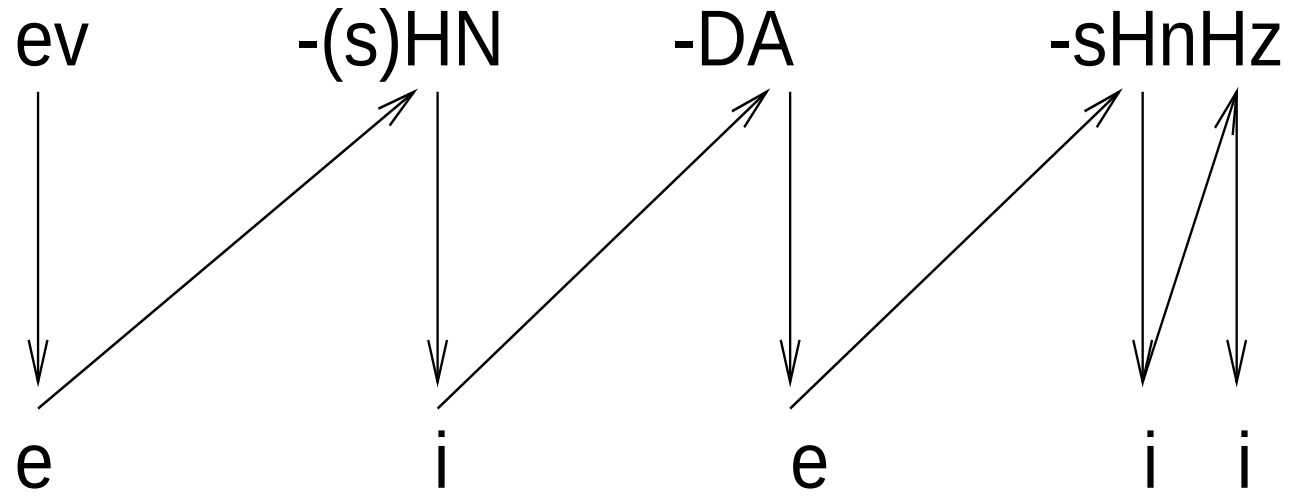

Figure V.1: Progressive nature of vowel harmony

Below, some instances demonstrating the realization of two-level rules for the vowel harmony are presented. Morphological decomposition is given on the left whereas the morphophonemic process is shown on the right.

```
(18) a. saat  -ler -imiz   -den          LR: saát-lAr-(H)mHz-DAn
        watch -PLU -POSS1p -ABL          SR: saat0ler00i0miz0den
        'from our watches'

     b. alkol   -süz -ler                LR: alkól-sHz-lAr
        alcohol -DER -PLU                SR: alkol0süz0ler
        'the ones without alcohol'

     c. karşı    -da  -ki  -ler -e       LR: karşı-DA-ki-lAr-(y)A
        opposite -LOC -REL -PLU -DAT     SR: karşı0da0ki0ler0000e
        'to the ones opposite'
```

### V.1.2 Consonant harmony

Recall from Subsection IV.2.2 that in principle consonant clusters in Turkish are claimed to be either voiced or voiceless. Unfortunately, things are more complicated than this. Processes involving consonant harmony (assimilation) in Turkish can be grouped into two main categories. a) devoicing/voicing of plosive consonants in the root-final or in the suffix-final position; and b) devoicing/voicing of the initial plosive of certain suffixes.

#### Final Stop Devoicing/Voicing

The first one of the rules concerning consonant assimilation is the devoicing of the word-final or suffix-final stops. Generally speaking, Turkish nominal or verbal roots do not end with the voiced stops (e.g., `b`, `c`, `d`, `g`). However, some borrowed roots such as (e.g., `Ahme`*d*, `ren`*g*, `ahen`*g*, `kita`*b*, `mektu`*b*, `ila`*c*) have a final voiced stop. The final stops in such roots often alternate to their voiceless correspondents when the next segment in a suffix is consonantal.

Although Turkish phonological system forces all the voiced word-final stops to become voiceless, there are cases where the final-stop devoicing does not take place. Some of these are `ad` ‘name’, and `hac` ‘pilgrimage’ . These require special treatment, as their word-final consonant must not get voiceless violating the rule. For instance,

```
(19) a. hac → hac
        pilgrimage-NOM
        ‘pilgrimage’

     b. ad-DAn → addan
        name-ABL
        ‘from the name’
```

In addition, there are a few borrowed words in which the original cluster contained as first member a voiced stop and as second member a voiceless consonant, a combination nonexistent in Turkish. In these few cases the first stop is devoiced when the epenthetic vowel drops. Thus:

(20) za*b*ıt-A → za*p*ta
seizure-DAT
‘to the seizure’

In the same way that the devoicing occurs when a voiced morpheme-final consonant is affected by a suffix-initial non-vowel, a voiceless consonant may become voiced when assimilated by the initial vowel of the following suffix. However this voicing process is not a regular one. Although it is commonly observed in polysyllabic words, it can also be seen in monosyllabic words. This process also varies in effectiveness from one voiceless stop to another. As Schaaik ([30]:122-124) reports the final stop devoicing rule notably exhibits ”a rather skewed distribution of regular words and exceptions”. He also adds that in some cases exceptionality seems to be the dominant property of a certain word class. Therefore, the rules for each morpheme-final stop (i.e., `ç`,`k`,`t`,and `p`) should be handled separately. Note that since the final-stop devoicing/voicing process is not regular and changes for both native and borrowed words, multiple surface forms for the same lexical form can be observed even in the written language. Here are some examples from ([2]:94-95).

(21) `sütü or südü, kulpu or kulbu, layıkı or layığı`

To avoid such multiple occurences and establish a standard, [34] is taken as the ultimate reference for the lexical entries.

**Final Voicing/Devoicing of `ç`** : `ç` at the final position in the monosyllabic words do not usually become voiced when a suffix with an initial vowel is appended. `saç` 'hair', `haç` 'crufix', `kaç` 'escape' are examples to these word forms. However, there are a few number of contradicting word forms such as `uç`, `güç`. On the other hand the word-final `ç` in polysyllabic words almost always become voiced. There are also exceptions to this condition. Most of these exceptions are compound word forms such as `soliç`, `verkaç` and `vazgeç`. The preliminary statistical work provided by Schaaik [30] strenghtens the claim that the words ending in `ç` is very sensitive to final stop voicing. In monosyllabic words ending in a consonant cluster which contains a dental-alveolar consonant {`l`,`n`,`r`} as the first consonant and `ç` as the second one, that word-final `ç` gets voiced (e.g., `genç`-A → `gence` 'to the young'). Once more there are, though very little in number, exceptions for this condition such as `linç` and `sürç` where voicing process does not take place.

Considering the non-harmonic distribution of the exceptional cases, the word/morpheme final voicing for `ç` is designed and implemented as follows:

- The ç at the word/morpheme-final position of all polysyllabic entries and at the word-final clusters nç, rç, lç, vç of all entries (both monosyllabic and polysyllabic) becomes voiced if the initial phoneme of the following suffix is a vowel.

- The final devoicing is applied for the roots ending with a c. Therefore, the c at the final position of the word forms, especially the borrowed ones, corresponds to ç (e.g., ilac will have the surface form ilaç.[3] The word-final c also designates the existence of voicing. That is, monosyllabic words such as uç 'tip', öç 'revenge' for which voicing is enabled violating the principle above will have the lexical forms uc and öc correspondingly. The word hac 'pilgrimage' which is represented with the same form in the surface level, however, seems to be an exception. The problem is removed by assigning the lexical representation hacc (for the gemination process) to the word.

- Finally, the irregular word forms that end with a word-final ç but do not become voiced, will be designated by a C. This meta-phoneme at the word/morpheme-final position will correspond to ç. Note that these irregular forms such as ölç 'measure', vazgeç 'abandon' are few.

(22) a. öc → öç
revenge-NOM
'revenge'

b. öc-(s)HN → öcü
revenge-POSS
'his/her revenge'

c. öc-(y)lA → öçle
revenge-INS
'with revenge'

[3] Note that the lexical form can also be considered as ilaç.

d. `ağaç-(y)H` → `ağacı`
`tree-ACC`
`‘tree (object)’`

e. `hınç-(y)A` → `hınca`
`grudge-DAT`
`‘to the grudge’`

**Final Devoicing/Voicing of `b/p`** : Devoicing/voicing of the word-final `b/p` pair also exhibits the same characteristics. There are again exceptional words such as `ip` ‘rope’ and `miyop` ‘myopia’ that violate the general tendency: The word-final `p` becomes voiced in the context where the next segment is a vowel. These exceptional words are mostly borrowed or mono-syllabic like `ip` ‘rope’, `sap` ‘handle’, `yap` ‘do’, or compounds such as `yarıçap` ‘radius’. Schaaik [30] also mentions in his preliminary work that 75% of the nouns ending in `p` are sensitive to final-stop voicing. Yet, 25% is a great ratio to enumarate the exceptional cases. Thus, devoicing/voicing of the final `b/p` is formulated as follows:

- A word-final `p` never becomes voiced. Therefore, both native and borrowed forms such as `ip`, `yarıçap` that have a voiceless stop that never becomes voiced will have word-final (actually lexeme-final) `p` in their lexical representations.

- A word-final `b` becomes unvoiced if the lexeme takes no suffix or takes a suffix with a morpheme-initial non-vowel.

- For a very limited number of borrowed words such as `ab` ‘water’ and `lebaleb` ‘fully’, the meta-phoneme `B` is introduced to formulate the case where the lexeme-final `b` is not voiced (e.g., `lebaleB` → `lebaleb`).

(23) a. `ip-(y)H` → `ipi`
`rope-ACC`
`‘to the rope’`

b. `yap-(H)yor` → `yapıyor`
`do-CONT`
`‘he/she is doing’`

c. `dolab` → `dolap`
`wardrope-NOM`
`‘wardrope’`

d. `cevab-DAn` → `cevaptan`
`answer-ABL`
`‘from the answer’`

**Final Devoicing/Voicing of `d/t`** : According to Schaaik [30], the least sensitive group of words to final stop voicing are the ones ending in a `t`. He claims that only 690 out of 1835 nouns ending in `t` are subject to that rule. In monosyllabic words that end in `t`, voicing does not generally take place whereas in polysyllabic ones, `t` usually becomes voiced (i.e., `t` corresponds to `d`). For instance,

(24) a. `at-(y)H` → `atı`
`horse-ACC`
`‘horse (object)’`

b. `sat-(H)yor` → `satıyor`
`sell-CONT`
`‘he/she is selling’`

c. `kanat-(y)A` → kana*d*a
`wing-DAT`
`‘to the wing’`

d. `kilit-(H)m` → kili*d*im
`lock-POSS`
`‘my lock’`

In any case, voicing for `t` can not be formulatted to cover a large set of words since there are many exceptional cases such as `sepet`, `adalet`, `millet`, `artist`, `bulut`, `izmarit`, `dinamit`, `linyit` and etc. Besides there are many monosyllabic words that have a final `t` which becomes voiced in proper contexts. These also make up a large set of exceptions. Thus devoicing/voicing of word/morphme-final `t` is described as follows:

- Lexical (native or borrowed) entries are considered to end in a `t` or `d` or `D`. Whether a word is mono-syllabic and/or borrowed does not have any effect in the application of voicing/devoicing rules for `t` and `d`.

- Word-final voicing is not applicable for the words that end in a `t`. That is, if a word ends in a `t`, that `t` does not change in any context.

- If a word ends in a `d`, then that `d` becomes a `t` when the word does not take any suffix or is suffixed by non-vowel initial morpheme. Otherwise, i.e., when the word takes a vowel initial suffix, `d` does not undergo any change.

- There is, however, a limited set of words such as `ad`, `öd`, `kod`, `mod` that violates the rule above. That is, the final `d` in these words is never voiced. These are supposed to have a word-final meta-phoneme `D` in their lexical forms (e.g., `aD` → `ad`).

```
(25) a. sepet-(H)m → sepetim
        basket-POSS
        'my basket'

     b. kanad → kanat
        wing-NOM
        'wing'

     c. aD → ad
        name-NOM
        'isim'

     d. aD-DAn → addan
        name-ABL
        'from the name'
```

e. `fark-ed-DH` → `farketti`
`difference-AUX-PAST`
`‘it differed’`

f. `gid-(H)yor` → `gidiyor`
`go-CONT`
`‘he/she is going’`

g. `gid-DH` → `gitti`
`go-PAST`
`‘he/she went’`

**The k/ğ, k/g, g/ğ alternations** : The word-final k in monosyllabic words is rarely voiced. Some of these exceptional cases are gök ‘sky’, çok ‘many’, and yok ‘non-existent’. In polysyllabic words and all of those words that end in a nk cluster, the word final k alternates to ğ or g respectively. These alternations are modelled as follows:

- The word/morpheme final k in polysyllabic words corresponds to the semi-glide ğ when surrounded by a vowel context. There are exceptional words such as hukuk ‘law’, ahlak ‘moral qualities’, and ittifak ‘alliance’ to this rule. These are marked by a final q (Such borrowed words, mostly of Arabic, actually end in a q).

- The word-final k preceded by n changes to g. Note that words like aheng, reng, and etc. are represented lexically as ahenk, renk, and etc. That is, the final g does not alternate with k but vice versa. The main reason for this is the existence of borrowed words such as bumerang, brifing, and şezlong where the final g never becomes devoiced. Some exceptional words such as bank, tank, link still exist. These are represented lexically by a word-final q (e.g., tanq, linq, etc.).

- The word-final `k` surrounded by vowels in some monosyllabic native words alternates with `ğ`. Such words make up a very limited set of words such as `gök`, `çok`, and `yok`. These are marked by the meta-phoneme `K`. Thus these exceptional words are represented by a word-final `K` lexically.

- The word-final `g` preceded by `o` corresponds to `ğ` if the initial phoneme of the following suffix is a vowel.

(26) a. yata*k*-(y)A → yata*ğ*a
bed-DAT
'to the bed'

b. gö*K*-(y)A → gö*ğ*e
sky-DAT
'to the sky'

c. ahla*q* → ahla*k*
morility-NOM
'morality'

d. huku*q*-(y)A → huku*k*a
law-DAT
'to the law'

e. ren*k*-(y)H → ren*g*i
color-ACC
'color (object)'

f. tan*q*-(s)HN → tan*k*ı
tank-POSS
'is/her tank'

g. pedago*g*-(y)H → pedago*ğ*u
pedaqogue-ACC
'pedagogue (object)'

**Suffix-Inital Devoicing/Voicing**

There are two suffix initial meta-phonemes `D` and `C`. Suffix-initial voicing/devoicing is regularly applicable. Suppose that the suffix-initial dental consonants `d` and `t` is represented by the meta-phoneme `D`. `D` is resolved to a `d`, if the last phoneme in the stem is voiced; and otherwise to a `t`. `D:d` is supposed to be a default correspondence. Thus `D` is forced to become `t` (i.e., devoiced) in the proper context or left as `d`.

Recall that to handle the words that end in a `ç` but does not become voiced, `C:ç` is assigned as the default correspondence. Therefore, unlike `D`, `C` is forced to become `c` (i.e., voiced) in the appropriate context.

```
(27) a. sokak-DAn → sokaktan
        street-ABL
        'from the street'

     b. kaç-DH → kaçtı
        escape-PAST-PERS3s
        'he/she escaped'

     c. keman-CH → kemancı
        violin-DER
        'one who plays violin'

     d. saát-CH → saátçi
        watch-DER
        'one who repairs/sells watches'
```

The two-level rules making up the consonant harmony are given below. Table V.3, Table V.4, and Table V.5 show the two-level rules for final stop devoicing, suffix-initial stop and final stop voicing/devoicing respectively. Note that RCFSDV denotes the regular expression for the right context of final stop devoicing; Refer to the rules file in the appendix for a detailed description.

Table V.3: The word/morpheme-final voiced stops become voiceless

```
CsV:CsVless   ⇔       \[CsV:] _ (CsV:0-c:0) RCFSDV;
where         CsV      in (b c d)
              CsVless  in (p ç t)
matched;
```

Table V.4: The rules for alternations of suffix-initial meta-phonemes

```
D:t  ⇔   :CsVless (:0)* MB _;

C:c  ⇔   \:CsVless (:0)* MB _;
```

Table V.5: The word/morpheme-final voiceless stops get voiced

```
ç:c  ⇔   _ (:0 - FSM)* MB (:0)* :V;
         :V CsLatNas _ (FSM) (RB) MB (:0)* :V;

k:ğ  ⇔   :V _ (:0 - FSM)* MB (:0)* :V;

g:ğ  ⇔   o _ (:0 - FSM)* MB (:0)* :V;

k:g  ⇔   :V n _ :0* :V;
```

### V.1.3 Vowel epenthesis

Although vowel clusters are not existent in Turkish, some borrowed roots such as `saat` 'hour' and `vaad` 'promise' violate this principle. However, as a consequence of this principle vowels may drop at morpheme boundaries. Some instances for the vowel epenthesis are given below.

(28) a. `araba-(`*H*`)m` → `arabam`
`car-POSS1s`
`'my car'`

b. `ara-(`*H*`)n-mAk` → `aranmak`
`seek-REFX-DER`
`'to seek oneself'`

c. `bekle-(`*H*`)yor` → `bekliyor`
`escape-PAST-PERS3s`
`'he/she is waiting'`

Note that the first two instances exemplify the epenthesis of the optional suffix-initial vowel whereas the final one displays a special case. This process will be discussed in Subsection V.1.7. There is also an epenthetic vowel observed in some specific contexts (within the root forms). This process will be expressed as a part of borrowed-phonology although there is a small set of Turkish nouns, referring almost exclusively to body parts in which the vowel in the final syllable drop in certain contexts. Below, some instances of this process for some native roots are presented.

```
(29) a. bur𝑢n-(s)HN → burnu
        nose-POSS3s
        'his/her nose'

     b. om𝑢z-(y)A → omza
        shoulder-DAT
        'to shoulder'

     c. çev𝑖r-Hl → çevril
        surround-PASS
        'be surrounded'
```

**Suffix-Initial vowel epenthesis**

The suffix-initial vowels are represented by the meta-phonemes `A` and `H` in enclosing parentheses. Two examples of the suffixes with the morpheme-initial optional vowel segment are: `-(H)m` "the first person singular possesive suffix", `-(H)ş` "the suffix of reciprocal voice". When the preceding morpheme ends in a consonant, `A` and `H` are resolved as a vowel in accordance with the vowel harmony rule, otherwise they drop. Table V.6 illustrates the two-level rule for this process.

```
(30) a. oda-(𝐻)nHz → odanız
        room-POSS2p
        'your room'

     b. bekle-(𝐻)ş-DH-k → bekleştik
        wait-RECP-PAST-PERS
        'we altogether waited for'
```

Table V.6: The two-level rule for suffix-initial vowel epenthesis.

```
SIV:0  ⇔     :V (:0)* MB OPHL _ OPHR;
       where  SIV in (A H E);
```

### V.1.4 Consonant epenthesis

As a consequence of the fact that there is a limited number of consonant clusters that may appear in a word, the suffix initial consonants after an immediately preceding non-vowel at the morpheme boundary usually drop. These suffix-initial consonants are `n`, `s` and `y`. These consonants are optional. Actually their job is to coalesce the word-final vowel of the preceding morpheme with the suffix-initial vowel of the following morpheme. When the word-final phoneme is not a vowel, it is dropped. Note that these optional consonants which drop according to the environment are shown between two parentheses. Some of the suffixes with the initial optional consonant segment are: `-(y)H` "the suffix of accusative case", `-(s)HN` "the suffix of third person singular possesive", `-(n)Hn` "the suffix of genitive case". Table V.7 displays the two-level rule for the epenthesis of optional suffix-initial consonant.

(31) a. ev-(*n*)Hn → evin
house-GEN
'of the house'

b. ev-(*n*)Hn-(*y*)DH → evindi
house-GEN-AUX(past)
'it was of the house'

c. saát-(*y*)A → saáte
watch-DAT
'to the watch'

Table V.7: The two-level rule for suffix-initial consonant drop.

```
SIC:0  ⇔     :Cs+ (:0-[N: | Y:])* (FSM) MB OPHL _ OPHR;
       where  SIC in (n s y);
```

There are some other processes in which consonants may drop. These are mostly specific to a closed set of suffixes and will be mentioned in Section V.1.7.

### V.1.5 Reduplication

Reduplication is a phonological process. At first sight, this may seem like a prefix but it is not actually a morphophonemic phenomenon. In any case, a reduplication process occurs for a closed set of words. These are some adverbs and adjectives, especially adverbs of manner and adjectives of color. This reduplication process intensifies the meaning of the word. Some examples are presented below.

(32) a. *kıp*-kırmızı
*RUP*-red
'very red'

b. *yem*-yeşil
*RUP*-green
'very green'

c. *çar*-çabuk
*RUP*-quickly
'very quickly'

The reduplication process possesses some degree of generality for a category of words. We examined almost all the lexicalized instances of words for which the reduplication is applicable and concluded that some kind of regularity exists. However, we must admit that it would be ad hoc formalization. We tabulated the relations among phonemes for the reduplication process and formulated the process as two-level rules accordingly. Design issues can be summarized as follows:

- A dummy prefix, namely `RUP-`, is introduced. Note that `RUP-` consists of three consecutive archiphonemes `R`, `U`, and `P` respectively. This is because of the fact that the reduplicated segment is a cluster of maximum three (minimum two) phonemes.

- Two-level rules that perform reduplication define nothing but the special correspondences for the archiphonemes `R`, `U` and `P`. These correspondences can roughly be described as follows:

  – If the initial phoneme of the reduplicated segment is a vowel then `R` corresponds to `0` (i.e., the (null) character symbol). For instance, `RUP-açık` 'clear' → `apaçık` 'very clear', `RUP-ayrı` 'different' → `apayrı` 'very different', and `RUP-ak` 'white' → `apak` 'too white'. If the initial phoneme of the word to which the reduplication is applied is a consonant then R always corresponds to that consonant. For example, `RUP-`*k*`ara` 'black' → *k*`apkara` 'too black'.

  – `U` always corresponds to the vowel of the segment which is reduplicated. This is the core of the reduplicated first syllable. As can be seen from the examples given above, reduplication process is actually applied to the initial segment of words and the first part of this segment is the first syllable.

  – `P` corresponds to a consonant according to a predefined relation between the word-initial consonant and the following consonant (i.e., the initial consonant, the onset of the following syllable). For instance, when the first consonant and the following consonant of the segment are `k` and `r` respectively, `P` corresponds to `p`. Note that `P` corresponds to one of the four consonants:`p`, `s`, `m`, or `r`. We have searched for the lexicalized instances of the reduplication process from [34] and found that `P` is mostly mapped to `p` and only to `p` when `R`:0 exists (e.g., *ap*`ak`, *ap*`açık`, *ap*`az`, *ep*`eski`, *ıp*`ıssız`).

Table V.8 displays the possible correspondences of `P` for 53 instances taken from [34]. Refer to the rules file in the appendix for the relevant two-level rules and their explanations. Note that the first row denotes possible instances for the onset of the second syllable, and the first column represents the possible realizations of `P`. Each cell in the table gives the number of occurrences for each onset-realization pair. For instance, the number 6 in the cell `r-p` denotes that there are 6 realizations of `P:p` when the onset of the second syllable is `r`.

Table V.8: Special correspondences of reduplication process.

| | b | c | ç | d | f | g | ğ | h | j | k | l | m | n | p | r | s | ş | t | v | y | z | |
|---|---|---|---|---|---|---|---|---|---|---|---|---|---|---|---|---|---|---|---|---|---|---|
| m | | 1 | | | | | | | | 3 | | | | | 1 | 1 | 4 | | | 1 | 2 | 13 |
| p | | 1 | 1 | | | | 1 | | | 1 | 2 | | 3 | | 6 | 1 | | | 1 | 3 | 4 | 24 |
| r | | | | | | | | | | | | 1 | | 1 | | | | | | | | 2 |
| s | | | | 1 | | | 1 | | | | 1 | 2 | | 1 | 1 | | | 4 | 2 | 1 | | 14 |

### V.1.6 Loan-word Phonology

Borrowed words of the lexicon exhibit some nonproductive regularities. These regularities are, for the most part, reflections of the assimilation of borrowed words. The processes involved are degemination of consonant clusters, insertion of vowel into the consonant clusters, and shortening of final-consonant clusters.

**Degemination**

While in Arabic geminate consonant clusters are permitted in final postion, these are usually reduced or eliminated in Turkish borrowings. For such cases, it seems appropriate to consider the base to be a geminate-final-consonant morpheme which reduces the cluster before the following consonant-initial morphemes at morpheme boundary or before word boundaries.

```
(33) a. redd    -(y)A  LR: redd-(y)A
        refusal -DAT   SR: redd0000e
        ‘to the refusal’

     b. redd    -DAn  LR: redd-DAn
        refusal -ABL  SR: ret00ten
        ‘from the refusal’

     c. hazz     -(H)m    LR: hazz-(H)m
        pleasure -POSS1s  SR: hazz00ı0m
        ‘my pleasure’

     d. hazz     -(y)lA  LR: hazz-(y)lA
        pleasure -INS    SR: haz0000la
        ‘with pleasure’

     e. hacc        -(y)A  LR: hacc-(y)A
        pilgrimage -DAT   SR: hacc0000a
        ‘to the pilgrimage’

     f. aff         -DAn  LR: aff-DAn
        ‘forgiving’ -ABL  SR: afftan
        ‘from forgiving’

     g. aff         -0    LR: aff-0
        ‘forgiving’ -NOM  SR: af000
        ‘forgiving’
```

Table V.9 shows the two-level rule for degemination process. Other rules especially the final stop devoicing rule is adjusted accordingly.

Table V.9: Geminated final stops of borrowed words drop in certain contexts.

| | |
|---|---|
| Cy:0 ⇔ | [Cx: - Cx:0] _ (FSM) (RB) [ # \| MB OPHL y:0 OPHR :Cs \| MB :Cs]; |
| where | Cx in (b c d f h k l m n r s t v y z) |
| | Cy in (b c d f h k l m n r s t v y z) |
| matched; | |

### Epenthetic Vowel

There exists a derived high vowel in the second syllable of some disyllabic Turkish borrowings. This high vowel is short and tends to drop if a suffix with an initial vowel is added to the root. Some examples of these borrowings are listed below.

(34)
a. şeh*i*r (< Persian şehr) 'city'
b. öm*ü*r (< Arabic ömr) 'life'
c. eh*i*l (< Arabic ehl) 'competent'
d. nák*i*t (< Arabic nákd) 'cash'
e. vák*i*t (< Arabic vákd) 'time'

Contrasting with the words above the derived vowel does not drop in some borrowings such as `havuz` (< Ar. hawz) and `tohum` (< Prs. tohm). Tekin [35] relates these exceptional cases with the fact that `/hm/`, `/vz/`, and `/nf/` do not exist in native words. That is, the phonological system prevents the epenthesis of high vowel. Banguoğlu [2], on the other hand, explains this situation with the acceptability of these words by native speakers.

Tekin ([35]:pp107-108) formulates the process of the high vowel epenthesis in native bases by expressing the phonetic constraints as follows:

- The high vowel in the second syllable of disyllabic words that are names for body parts and are made up of an initial syllable ending in a vowel (i.e., in the form of V or CV), final syllable in the form `ğHr`, `ğHn`, `ğ-z`, `ğ-s`, `yHn`, `lHn`, `rHn`, `nHz`, `mHz`, and `vHç` drops when these words take a suffix with an initial (optional) vowel. There are a few native Turkish words such as `oğul` 'son' and `kayın` 'the brother in law' which are not the names of body parts but in which the high vowel of the second syllable drops in the same condition as above. In Example (35) the instances of this specification are given.

- Similarly, the high vowels in the second syllable of the verbal bases with a first syllable ending in a vowel and a final suffix in the form `yHr`, `vHr`, `ğHr`, `kHr`, `pHr`, and `kHl` drop when these verbal bases take a voice marker such as the `-(H)ş`, `-(H)n`, and `-(H)l` suffixes beginning with a (optional) vowel. Some examples of these verbal bases are: `ayır` 'part', `buyur` 'order', `sıyır` 'peel off', `kayır` 'support/back', `çevir` 'surround', `devir` 'knock down', `kıvır` 'curl', `savur` 'brag', `süpür` 'sweep', and `tükür` 'spit out'.

| | | | | | |
|---|---|---|---|---|---|
| (35) | a. bağ*ı*r-(H)m | → bağrım | bossom-POSS1s | 'my bossom' |
| | b. ağ*ı*z-(y)A | → ağza | mouth-DAT | 'to the mouth' |
| | c. göğ*ü*s-(H)n | → göğsü | breast-POSS2s-ACC | 'your breast' |
| | d. bey*i*n-(s)HN | → beyni | brain-POSS3s | 'his/her brain' |
| | e. om*u*z-(H)n | → omzun | shoulder-POSS2s | 'your shoulder' |
| | f. av*u*ç-(s)HN-(y)A | → avcuna | palm-POSS3s-DAT | 'to his palm' |

There are other exceptional borrowing words in which the high vowel in the second syllable does not drop. For instance the high vowel in the words `ahır` (< axu:r Persian) 'stable' and `havuç` (< havi:ç Persian) 'carrot' do not disappear when these words take suffixes with an initial vowel, although there are words such as `kahır` (< Ar. kahr) 'anxiety' and `avuç` 'palm' which have almost the same surface representation but in which the high vowel drops. The reason for that is actually can be grasped from the original forms of these borrowings: The high vowel is indeed not short but long. This explanation is sufficient enough to explain the cases `koyun` 'sheep' and `kayın` 'a kind of tree' whose surface represenations are exactly the same as the words `koyun` 'bossom' and `kayın` 'the brother in law' but the high vowel does not disappear in the first case but does so in the latter case. In other words, the high vowel in the second syllable of these words are indeed long. In any case, we have formulated this process in a very straight forward fashion. That is a high vowel in the second syllable of word bases with the form `CVCHC` or `VCHC` is allowed to drop when the preceding suffix (if exists at all) begins with a (optional) vowel. Table V.10 displays the corresponding two-level rules for this process.

Table V.10: The high vowel epenthesis in word bases.

```
H:0  ⇔   V FSM Cs _ :Cs (RB) MB [OPCD :V | OPNV | :V];
```

### V.1.7 Morpheme specific processes

In this section, the morpheme specific rules will be presented. These rules are very far from being general and designed to implement exceptional cases.

**The aorist suffix `-(E)r` :**

One of the verbal suffixes in Turkish whose allomorphs are not entirely predictable from the phonemic context is the Aorist suffix. Many books describe two main morphemes for the Aorist suffix: `-(H)r` and `-Ar`. These also have allomorphs as a consequence of vowel harmony rules.

However this kind of a representation for the Aorist suffix puts a burden on the morphotactical design. First, one would have to classify the verbal stems into two: the ones taking the aorist `-(H)r` and others taking `-Ar`. This also requires some manipulation in the lexicon. For instance, the verbal bases might have an additional feature marking which one of the aorists that the base would take. Thus a great amount of processing will take place. Oflazer also [27, 28] follows the same way around in the computerized implementation. Depending on the description provided by [8], we have come up with a more compact and effective design that will remove the ineffciencies and disallow the redundancy present in [27, 28]. This design strategy introduces only a single aorist suffix `-(E)r` which will correspond to an appropriate allomorph in the related context. The description that forms the basis of our design policy for the formulation of the aorist suffix is as follows:

- All the polysyllabic verbal roots/stems and also the twelve monosyllabic verbal roots that end in a `l` or `r` take the aorist suffix `-(H)r`. These thirteen exceptional verbal roots are `al`, `öl`, `ol`, `gel`, `kal`, `bul`, `bil`, `var`, `ver`, `vur`, `gör`, `san`, and `dur`. In our design, `-(E)r` corresponds to `-(H)r` in this context.

- Apart from those thirteen roots, monosyllabic verbal roots take the aorist suffix `-(A)r`, that is `-(E)r` maps to `-(A)r`.

- If a verbal root ends in a vowel then the optional archiphoneme `E` drops, thus the allomorph of the aorist is `-r`. Thus the aorist `-(E)r` will have one of the following surface representations (i.e., allomorphs): `-ar`, `-er`, `-ır`, `-ir`, `-ur`, `-ür`, and `-r` depending on the context.

Note that the two-level rules for the implementation of the aorist suffix are nothing but versions of the vowel harmony rules and the suffix initial vowel epenthesis rule. The enumerated exceptional verbal roots are embedded into the rule file. Refer to the rules file in the appendix for a complete enumaration.

There is, yet another, morphophonemic process related with the aorist. The negative aorist suffix `-z` corresponds to (null) when it is followed by the first person singular or plural suffix. The corresponding two-level rule is given in Table V.11.

Table V.11: The two-level rule for negative aorist suffix epenthesis.

```
z:0  ⇔   MB m A: MB _ MB OPHL [ y:  H: OPHR m | y:  OPHR H: z];
```

**Dative Pronoun:**

The first and second person singular personal pronouns have each an irregular allomorph before the dative. That is, the base changes. Oflazer treats these forms as lexicalized items and puts them in the lexicon. This causes the first and second personal prounouns to be seperated from the other regular pronouns and there should be two distinct continuation classes for the prounouns in the lexicon file. We preferred another way, which we believe to be more effective and tractable from the view points of both implementation and morphological consistency. No allomorphs for the dative cases of `ben` 'I' and `sen` 'you' are placed in the lexicon, but, via a simple two-level rule, `e` in the base is mapped to `a` when these pronouns take the dative case suffix `-(y)A`. The two-level rule for handling this irregularity is given in Table V.12. The process of becoming +back for some forms of `ben` and `sen` can be explained by the backness effect of a velar (ngX$n$ (e.g., Te$n$ri > Tanrı). Therefore originating from the ancient Turkish, the following evolution can be foreseen:

`ben-ge sen-ge > ban-ga san-ga > bana sana`

Table V.12: The two-level rule for the exception of `e` to `a` alternation.

```
e:a  ⇔   [b|s] _ n FSM (RB) MB OPHL y:  OPHR A: [# | MB];
```

```
(36) a. ben -(y)A →  bana
        I   -DAT
        'to me'

     b. sen -(y)A →  sana
        you -DAT
        'to you'
```

**Morphemes ending in `N` :**

Some bound and free forms are marked by a final archiphoneme `N`. These are the third person possesive suffixes `-(s)HN` and `-LArHN`, the third person personal pronoun `o` 'he/she/it', the demonstrative pronouns `bu` 'this', `şu` 'that', and `o`, the relativization suffix `-ki` which makes adjectives and pronouns from nouns, and finally the reflexive pronoun `kendi` 'oneself'. This is because all these forms, when followed by the case suffixes, exhibit irregularites. Some instances of invalid surface forms and corresponding valid surface forms are given in (37).

| | | | | |
|---|---|---|---|---|
| (37) | a. ev-(s)H-DA | → *evide | ev-(s)HN-DA | → evinde |
| | | | house-POSS3s-LOC | → 'at his/her house' |
| | b. ev-LArH-(y)A | → *evleriye | ev-LArHN-(y)A | → evlerine |
| | | | house-POSS3p-DAT | → 'to their house' |
| | c. şu-(y)H | → *şuyu | şuN-(y)H | → şunu |
| | | | that-ACC | → 'that (object)' |
| | d. o-(y)H | → *onu | oN-(y)H | → onu |
| | | | he-ACC | → 'him' |

Oflazer [27, 28] solves this problem by introducing a second group of case suffixes having the same continuation class but different surface forms. Basically, the case suffixes in this second group have a suffix-initial `/n/`. This approach puts a lot of burden on the morphotactical design. Besides, the formation of a second case group causes a misleading impression that there are two case groups in Turkish. The phoneme `/n/` that is inserted between case suffixes and the word forms stated above is often called the *pronominal* `n`. It is pronominal since other pronouns such as the reflexive pronoun `kendi` also requires it before case suffixes. The pronominal `n` does not drop even if the following suffix (e.g., `-DAn`, `-DA`) begins with a consonant. In our approach, the meta-phoneme `N` is utilized to designate the pronominal `n`. `N` has the default correspondence (null), and the special correspondence `n` in the specified context. Thus irregularities are overcome depending on the underlying form. Refer to the rules file in the appendix for the corresponding two-level rule.

**The third person plural possesive suffix `-LArHN` :**

The "`-LAr`" segment of the third person plural possesive suffix `-LArHN` drops when it is preceded by the plural suffix `lAr`. Oflazer [27, 28] solves this problem in the morphotactic component by not allowing `-lArH` to follow `-lAr`. This creates redundancy in the morphotactics by duplicating certain states and transitions. We prefer a more natural way. We let "`-LAr`" drop after `-lAr`. For example,

```
(38)    e v - l A r - L A r H N - ( y ) A → evlerine
        e v 0 l e r 0 0 0 0 i n 0 0 0 0 e
        ‘to their house(s) / to his/her house(s)
```

The morphophonemic rules for these alternations are given in Table V.13. Refer to the rules 41, 42, 43, and 44 in the appendix for details.

Table V.13: The morphophonemic process for `-LArHN`.

```
L:0  ⇔   MB l A: r MB _ ;
A:0  ⇔   MB L:0 _ ;
r:0  ⇔   MB L:0 A:0 _ ;
```

**The word `suY` :**

The word `su` 'water' represents one of the important exceptions in Turkish. For instance,

```
(39)  a.  su-(s)HN-DA  →  *susunda   suyunda  'in his/her water'
      b.  su-(n)Hn     →  *sunun     suyun    'of the water'
      c.  su-DA                      suda     'in water'
```

The inflectional and derivational forms of `su` can be generated as a regular form on the basis of an underlying form. We assume that the lexical form `suY` underlies surface forms like `su` "`suY-NOM`" and `suyu` "`suY-(s)HN`". The two-level rule in Table V.14 displays the formulas for the morphophonemic processes related with the archiphoneme `Y`. Note that `Y` corresponds to `y` when it precedes a suffix-initial vowel. In the corresponding two-level rule, the contexts following the morpheme-boundary are critical: `OPNV` denotes the regular expression describing the context where the suffix-initial optional vowel does not drop, and `OPCD` denotes the context where the suffix-initial optional consonant drops.

Table V.14: Modelling the `suY` exception.

```
Y:y  ⇔   _ (FSM) (RB) MB [OPNV | :V | OPCD];
```

**The causative suffixes -(DH)X and -(D)HX :**

The causative suffix is a very irregular one. It has many allomorphs. We have modelled the morphophonemics for this suffix in accordance with Turkish morphotactics. It is assumed that -(DH)X follows the transitive verb bases and -(D)HX comes after the intransitive ones. Apart from the causative-transitive suffixes -(DH)X and -(D)HX, there are the causative-agentive and causative-intensive suffixes with the same surface form -(D)HX. There are three related two-level rules responsible from the appropriate correspondences.[4] These morphophonemic processes are formulated according to the rules stated in [[8],pp:43-44]. Let us give some examples to the realizations of these suffixes.

```
(40) a. anla       -(DH)X  -(D)HX  → anlattır
        understand -CAUS_T -CAUS_A
        'make someone explain something to someone else'

     b. bit    -(D)HX  -(D)HX  -(D)HX  → bitirttir
        finish -CAUS_T -CAUS_A -CAUS_I
        'have someone to make someone else to end something'

     c. böl    -(DH)X-mAk → böldürmek
        divide -CAUS_T    -DER(infinitive)
        to get something divided by the agency of someone
```

**The passive voice suffix -(H)L :**

The passive voice suffix, -(H)L has the same surface representation with the reflexive voice suffix, -(H)n when the preceding verb base ends in a vowel or /l/. Note that the default correspondence for /L/ is /l/. Refer to the rule 45 of the rules file in the appendix for details. Here are some examples.

```
(41) a. yıka -(H)L -DH → yıkandı
        wash -PASS -PAST
        'was washed'
```

[4] Refer to the rules 38, 39, and 40 of the rules file in the appendix.

```
b. böl    -(H)L -DH → bölündü
   divide -PASS -PAST
   'was divided'

b. yık      -(H)L -mHş → yıkılmış
   demolish -PASS -NARR
   'was demolished (reported speech)'
```

**Morphophonemics for Person-Number Suffixes :**

There are morphotactical variations for some verbal person-number suffixes. Instead of assigning two distinct morphemes such as `-(y)Hm` and `-m` for the first person singular suffixes for different morphotactic continuations, it is theoretically more sound and practically more efficient to introduce one single morpheme such as `-(yH)m` that has various allomorphs in morphotactical contexts. For instance, `-(yH)m`, `-(sH)n`, and `-(sH)nHz` correspond to `-m`, `-n`, and `-nHz` respectively when they are preceded by `-DH`, `-(y)DH`, `-mHş`, and `-(y)mHş`; whereas they correspond to `-(y)Hm`, `-sHn`, and `-sHnHz` after another group of tense-aspect suffixes. Refer to the rules 46, 47, and 48 of the rules file in the appendix for details. Some instances involving these correspondences are given below.

```
(42) a. gel  -DH   -(yH)m  → geldim
        come -PAST -PERS1s
        'I came'

     b. gel  -Hyor -(yH)m  → geliyorum
        come -CONT -PERS1s
        'I am coming'

     c. gel  -(y)AcAk -(y)Hz  → geleceğiz
        come -FUTR    -PERS1s
        'We will come'

     d. gel       -(y)Hn  → gelin
        come -IMP -PERS2p
        'come (2nd person plural)'
```

## V.2 Turkish Morphotactics

While expessing the morphotactical constraints, we have used some formulation techniques in [29] to make morphotactical statements in a concise and clear way. These formulations can be considered a step towards greater economy in comparison with finite-state diagrams. Still more economy might be achieved by setting up new cover symbols for recurrent elements.

### V.2.1 Nominal Paradigm

Nominal morphotactics of Turkish is quite complex. It applies to nouns and adjectives. It consists of inflections of noun bases, derivations from noun bases to various parts of speech.

#### Nominal inflections

The paradigm for noun inflections is relatively simple. The relative order of these suffixes can be observed from the chart in Figure V.2.

| $A$ | $B$ | $C$ | $D$ | $E$ |
|---|---|---|---|---|
| `SUBSTANTIVE BASE` | `Plu.` | `Poss.` | `Case` $D_1$ $D_2$ $D_3$ $D_4$ | `Rel.` |

Figure V.2: The relative order of substantive inflectional suffixes.

In this chart,

($A$) It represents the nominal base. A nominal base may be:

- an adjective or noun root

- a nominal stem

($B$) It denotes the number suffix `-lAr`, `Plu`.

($C$) It denotes the group of possesive suffixes, `Poss`. Possesive suffixes are: `-(H)m`, `-(H)n`, `-(s)HN`, `-(H)mHz`, `-(H)nHz`, `-LArHN`.

($D$) It represents the group of case suffixes, `Case`.

Case suffixes are: `-(y)H` 'accusative', `-(y)A` 'dative', `-DA` 'locative', `DAn` 'ablative' `-(n)Hn` `-(y)lA` 'instrumental'.[5]

Note that the case group is partitioned into three mutually exclusive subclasses according to the continuation class:

- Subclass $D_1$ contains the locative case suffix `-DA` and the genitive case suffix `-(n)Hn`; these are the only suffixes that precede the relativization suffix `-ki`.

- Subclass $D_2$ consists of the accusative case suffix `-(y)H` that does not take predicative suffixes (e.g., `Ankara'daymış` 'he was in Ankara' is valid whereas `*Ankara'yımış` is not valid). These suffixes also do not take the relativization suffix `-ki`. `-y(H)` actually constitutes a terminating state in morphotactic description.

- Subclass $D_3$ consists of suffixes which do not take the relativization suffix `-ki` but the predicative suffixes.

- Subclass $D_4$ represents the nominative case which has the phonologically null surface form.

[5] Banguoğlu also considers `-cA`, `-lH`, `-sHz` (equative, minitive and private, respectively) as cases; see ([2]:pp329-331).

($E$) It denotes the relativization suffix `-ki`. `-ki` is actually considered to be a derivational suffix that derives pronouns (e.g., `evinki` 'the thing that belongs to the house') or adjectives `evdeki` 'the thing at the house' from nominal bases. However, as the derivation takes place within the same paradigm, `-ki` is treated like an inflectional suffix.

As stated above predicative suffixes can be attached onto nominal bases to form verbal and adverbial constructions. These suffixes consist of a group of tense/aspect and person-number suffixes. The chart displaying the relative orderings for predicative suffixes is given in Figure V.3. (43) shows examples of predication.

| $K$ | $L$ | $M$ | $N$ |
|---|---|---|---|
| Person<br>Number<br>Suffixes | Inter.<br>suffix | Tense<br>Aspect<br>Suffixes<br>Grp-2 | Person<br>Number<br>Suffixes |
| $K_1$<br>-lAr | -mH | $M_1$<br>-(y)mHş | $N_1$<br>-(yH)m<br>-(y)Hz<br>-(sH)n<br>-(sH)nHz |
| | | $M_2$<br>-(y)DH | |
| | | $M_3$<br>-(y)sA | $N_3$<br>-(yH)m<br>-k<br>-(sH)n<br>-(sH)nHz |
| | | $M_4$<br>-DHr | |
| | | | |

Figure V.3: The relative order of predicative suffixes.

```
(43) a. hasta -DHr
        ill   -COP
        's/he is ill'

     b. ev -DA  -DHr -lAr  → evdedirler
        ev -LOC -COP -PERS
        'They are (certainly) at home'
```

```
c. hasta -(yH)m → hastayım
   ill   -PERS
   'I am ill'

d. arkadaş -(s)HN  -(y)lA -(y)mHş  -(sH)n → arkadaşıylaymışsın
   friend  -POSS3s -INS   -AUX_INF -PERS
   'You were reportedly with his/her friend'
```

The valid morphotactic sequences will be presented in segments to facilitate more clear and modular formulations. Firstly, the nominal inflections will be treated.

- *Abc*[6]

  The possible combinations of word forms that can be retrieved from this sequence are:

```
(44) a. ev     A
        house
        'house'

     b. ev -lAr → evler Ab
        ev -PLU
        'houses'

     c. ev -(H)m   → evim Ac
        ev -POSS1s
        'my house'

     d. ev -lAr -(H)m   → evlerim Abc
        ev -PLU -POSS1s
        'my houses'

     e. ev -lAr -LArHN  → evleri Abc
        ev -PLU -POSS3p
        'their houses'
```

- *AbcD*

  where $D = (D_1 + D_2 + D_3 + D_4)$

  The possible combinations of word forms that can be retrieved from this sequence are:

[6] The lower case letters indicate the optionality.

```
(45) a. ev    -DA  → evde AD1
        house -LOC
        'at house'
     b. ev -lAr -(H)mHz -(y)H → evlerimizi ABCD2
        ev -PLU -POSS1p -ACC
        'to our houses'
     c. ev -(s)HN  -(y)A → evine ACD3
        ev -POSS3s -DAT
        'to his/her house'
```

- The continuation sequence for $E$ is: $EbD$

  That means the possesive suffixes are not attached to relativized nominals.

  Note that $E$ comes only after $D_1$, and this creates a recursive structure: $D_1$ $EBD_1$. For instance,

```
(46) a. ev    -DA  -kiN -(n)Hn -kiN -(y)A → evdekininkine
        house -LOC -REL -GEN   -REL -DAT
        'to the one who is related with the one at the house'
```

So, nominal inflections can be formulated as follows:

- $F_{Ni} = \boxed{A\ (bc(D_1e + D_2 + D_3 + D_4))}$

Now, let us examine the ordering relations among predicative suffixes. First of all, the person-number suffixes can be directly attached to the noun bases without taking a tense-aspect marker. In doing so, predication is supposed to be made for the present tense.

```
(47) a. iyi  -(y)Hz → iyiyiz
        fine -PERS
        'we are fine'
     b. okul   -DA  -(sH)nHz -DHr → okuldasınızdır
        school -LOC -PERS    -COP
        'You are (certainly) at the school'
```

When there are no case suffixes in the sequence, a long distance dependency may occur for the plural nominals. For instance,

```
(48)    kız  -lAr -(H)m   -(sH)n  → *kızlarımsın
        girl -PLU -POSS1s -PERS2s
        *‘you (PERS2s) are my girls’
```

The interrogative (question) suffix `-mi` comes before all the person-number suffixes (`except -lAr`), and the predicative negation is performed by introducing the word `değil` ‘not’ just after the nominal as a separate form[7]. For instance,

```
(49) a. ev    -DA  -mH   -(sH)n  → evde misin?
        house -LOC -QUES -PERS2s
        ‘Are you at home?’

     b. ev    -DA  -lAr    -mH   → evdeler mi?
        house -LOC -PERS3p -QUES
        ‘Are they at home?

     c. ev    -DA   değil -(yH)m  → evde değilim
        house -LOC  değil -PERS1s
        ‘I am not at home’
```

The present tense of nominal predication can be formulated as:

- $F_{Npp} = \boxed{A\ (F_{Npp1} + F_{Npp2})}$

  where

  - $F_{Npp1} = \boxed{(Bc\ \ F_{Npp1a}\ \ +\ \ c\ \ F_{Npp1b})}$
  - $F_{Npp1a} = \boxed{(lN_{1p} + N_{1p}M_4 + lM_4 + L)}$
  - $F_{Npp1b} = \boxed{(lN_1 + N_1M_4 + lM_4k_1 + Lk_1)}$
  - $F_{Npp2} = \boxed{bc(D_1 + D_3)(F_{Npp2a}\ \ +\ \ F_{Npp2b})}$
  - $F_{Npp2a} = \boxed{(lN_1 + N_1M_4 + K_1lm_4 + lM_4k_1 + Lk_1)}$
  - $F_{Npp2b} = \boxed{(F_{Npp1b} + K_1lm_4)}$

[7] Phrases are out of the scope of this work.

There is no long distance dependency involved related with person-number agreement between the number feature of the nominal stem and the person-number suffix for past, inferential, and conditional auxiliaries. Therefore, it is easier to formulate them.

The formula for the (narrative/inferential) past tense of nominal predication is given below:

- $F_{Nip} = \boxed{A\ (F_{Nip1} + F_{Nip2})}$

  where

  - $F_{Nip1} = \boxed{(Bc(lM_1 n_{1p}) + c(LM_1N_1 + lM_1k_1))}$
  - $F_{Nip2} = \boxed{bc(D_1 + D_3)(LM_1N_1 + lM_1k_1 + K_1lM_1)}$

Similarly, the (anterior) past tense of nominal predication can be formulated as:

- $F_{Nap} = \boxed{A\ (F_{Nap1} + F_{Nap2})}$

  where

  - $F_{Nap1} = \boxed{(Bc\ F_{Nap1a} + c\ F_{Nap1b})}$
  - $F_{Nap1a} = \boxed{(lM_2n_{3p})}$
  - $F_{Nap1b} = \boxed{(lM_2N_3 + lM_2k_1)}$
  - $F_{Nap2} = \boxed{bc(D_1 + D_3)(F_{Nap1b} + K_1lM_2)}$

Finally, the formula for the conditional aspect of nominal predication is as follows:

- $F_{Ncp} = \boxed{A\ (F_{Ncp1} + F_{Ncp2})}$

  where

  - $F_{Ncp1} \ = \boxed{(Bc\ \ F_{Ncp1a}\ \ +\ \ c\ \ F_{Ncp1b})}$
  - $F_{Ncp1a} \ = \boxed{(M_3 n_{3p} l)}$
  - $F_{Ncp1b} \ = \boxed{(M_3 N_3 l + M_3 k_1 l)}$
  - $F_{Ncp2} \ = \boxed{bc(D_1 + D_3)(F_{Ncp1b}}$

All of the the formulas for predicative suffixation is combined as follows:

- $F_{Np} = \boxed{F_{Npp}+F_{Nip}+F_{Ncp}\ +\ F_{Nap}}$

Note that, because of the high number of long distance dependencies involved, a considerable amount of duplication is observed in morphotactic encoding. Each "+" in a formulation actually introduces duplication of states and transitions. The overall formula for nominal inflections and predications is represented as:

- $F_N = \boxed{F_{Ni}+F_{Np}}$

### V.2.2 Verbal Paradigm

The morphotactics of verbal suffixes, their possible combinations, and restrictions on these combinations will be discussed in this section. Verbal inflections for finite verbs consist of a verb stem followed as a minimum requirement by a participle or tense suffix and a person suffix. A verb stem may be of three types:

1. It may consist of a single morpheme which is called a *verb root*. Note that in Turkish verb roots are usually monosyllabic. There are some roots which may receive either verbal or nominal suffixation: `ekşi` 'sour' or 'get sour' (adjective or verb respectively), `boya` 'paint' (noun or verb respectively). These are listed seperately in the lexicon for each syntactic category. Some examples of Turkish roots are listed below.

(50) `gel` 'come' `iç` 'drink' `yürü` 'walk' `giy` 'wear'

2. It may be the combinations of a substantive root and a denominal verb derivational suffix. This is called a `verb base`:

   (51) `taş` 'stone' (noun) `hasta` 'ill' (adjective)
   `taş-la` 'stone' (verb) `hasta-lan` 'become ill' (verb)

   Note that in Turkish verbalization and substantivization may also operate alternately and more than once: `çöz` (verb) 'untie,solve' → `çöz-üm` (noun) 'solution' → `çözüm-le` (verb) 'bring to a solution', `çözümle-me` (noun) 'bringing to a solution'.

3. It may include a verb root or a verb base plus one or more deverbal derivational suffixes. These derivational suffixes also called *voice* '`çatı`' suffixes. Some examples are given below:

   (52)

| | |
|---|---|
| `gör` | 'see' |
| `gör-üş` | 'converse, meet (lit. to see each other)' |
| `gö-rüş-tür-mek` | 'to make someone to see someone else' |
| `gö-rüş-tür-ül-mek` | 'to be made to see someone else' |

### Deverbal derivational suffixes

The first group of deverbal derivational suffixes consists of the so-called voice suffixes. The relative order for these suffixes is given in the chart in Figure V.4.

The ordering restrictions among the voice suffixes can be formulated as:[8]

[8] Note that orders with optional sequences are indicated by lowercase letters.

<table><tr><th>A</th><th>B</th><th>C</th><th>D</th><th>E</th><th>F</th></tr>
<tr><td rowspan="2">$A_1$<br>VERB<br>BASE<br>$A_2$</td><td>$B_1$<br>-(H)ş<br>recp.</td><td>$C_1$<br>-(DH)X<br>caus.<br>trans.</td><td rowspan="2">-(D)HX<br>caus.<br>agen.</td><td rowspan="2">-(D)HX<br>caus.<br>inten.</td><td>$F_1$<br>-(H)L<br>pass.</td></tr>
<tr><td>$B_2$<br>-(H)n<br>refl.</td><td>$C_2$<br>-(D)HX<br>trans2.</td><td>$F_2$<br>-(H)nHl<br>emph.</td></tr></table>

Figure V.4: The relative order of voice suffixes.

- $F_{D1a}$= $\boxed{A_1((B_1 + B_2)t_2f_1 + T_1f_1 + f_1 + f_2))}$
- $F_{D1b}$= $\boxed{A_2(B_1f_1 + b_1T_2f_1 + f_1 + f_2)}$

  where

  – $T_1 = C_1 + C_1D + C_1DE$
  – $T_2 = C_2 + C_2D + C_2DE$

- The overall formula is: $F_{D1}$= $\boxed{F_{D1a} + F_{D1b}}$

$(A)$ It represents the verb bases partioned into $A_1$ and $A_2$ since in our design verb bases are categorized mainly into two groups: Transitive verb bases and intransitive verb bases. Let us state briefly our convention of transitivity and intransitivity. Transitive verbs are those which take objects as explicit arguments, and intransitve verbs are those which do not take objects as arguments since the object of the action or condition is actually the subject. There is also ditransitive verb bases but these constitute a superset of transitive group and hence have no significance from the point of morphology. As stated above verb bases may be followed by a group of voice suffixes. Let us give a brief explanation about each of these suffixes.

($B$) This class denotes two mutually exclusive suffixes.

($B_1$) `-(H)ş` is called 'reciprocal'. It introduces the meaning of co-operative action to the verb base. That is, the action is done by more than one actor either in co-operation or in opposition. Sometimes the meaning changes may be observed as a consequence of the derivation. This suffix can be appended to both transitive and intransitive verb bases.

($B_2$) `-(H)n` is called the 'reflexive'. It enforces the meaning of "by one's self" or "on one's self" to verb, this can not be appended to intransitive verb bases.

($C$) This is the 'causitive-transitive suffix'. It gives the meaning of "cause to .." and "make someone do something" to the verb base. In doing so, it makes the verb transitive. There are many allomorphs of this morpheme. Allomorphic distribution is representable partly in terms of phonemic environment and partly in terms of continuation class. The morphophonemics of the causative suffixes are given in Section V.1.7. There are two variations of the causative-transitive suffix according to the continuation class: ($C_1$) and ($C_2$).

($C_1$) is `-(DH)X` and is appended to transitive verb bases.

($C_2$) is `-(D)HX` and is appended to intransitive verb bases.

($D$) It represents the 'causative-agentive' suffix. It introduces the meaning of "have something done through the agency of someone" to the verb base. An agent marked with the dative suffix `-(y)A` and a direct object are involved. The agent is the performer of the action. This suffix may be added to all transitive verb stems including those transitivized in (C).

(*E*) It denotes the 'causative-intensive' suffix with the meaning "get someone to do something through the agency of someone else". The causative meaning is intensified because of the repeated occurence of formally similar suffixes. Forms where causative-intensive suffix is used more than once are also heard occasionally. Both (D) and (E) are represented by the same morpheme `-(D)HX` which corresponds to various allomorphs depending on the phonemic context.

(*F*) $F_1$ represents the 'passive voice' suffix `-(H)L`.[9] Note that because of the context dependent morphophonemics, `-(H)L` occasionally corresponds to either -(H)l or `-(H)n` and thus the surface realizations of some verb bases in passive and reflexive voices are the same. An example to this is given in (53).

```
(53) a. koru   -(H)L       → korun
        defend -DER(PASS)
        'be defended by'

     b. koru   -(H)n       → korun
        defend -DER(REFL)
        'defend oneself'
```

$F_2$ represents the artificial passive voice suffix `-(H)nHl`. It is said to be artificial since it underlines the situation where the verbal stems that has already taken the passive voice allomorph `-(H)n`, are affixed by `-(H)l`. This is mainly because of the fact that the passive feature of `-(H)n` has not been completely accepted by native speakers. That is, the following `/Hl/` segment emphasizes the passive feature. Some examples of verbal stems that end with a vowel or `/l/` are `ye-n-il`, `de-n-il`, `hazırla-n-ıl`, and `bul-un-ul`. The example below illustrates the function of the artificial passive suffix.

[9] See Section V.1.7 for the morphophonemics of the passive voice suffix.

```
(54) a. koru   -(H)nHl      →   koru-(H)nHl
        defend -DER(PASS)       koru0000nul
        'be defended by'
```

The other group of deverbal verb suffixes that may be appended to the above group consists of the possibility suffix `-(y)A`, the verb predicative negation `-mA`, compound verb suffixes such as `-(y)Abil`, `-(y)Ayaz`, and `-(y)Hver`. The verbs derived by compound suffixes are called *descriptive verbs* by Banguoğlu ([2],pp:488-494). Figure V.5 shows the relative order of remaining deverbal verb suffixes.

| G | H | I |
|---|---|---|
| -(y)A<br>poss. | -mA<br>neg. | -(y)Abil<br>-(y)Adur<br>-(y)Hver<br>-(y)Agör |

Figure V.5: The order for other deverbal verb suffixes.

The formula $F_{D2}=\boxed{(h+gH)\ i}$ summarizes the relations among G,H, and I. Note that the possibility suffix `-(y)A` can not be used alone; it is always accompanied by the negation suffix.

The sum of all possible combinations from A to I provided with the above restrictions will give the deverbal verb stem construction: $F_D=\boxed{F_{D1}\ F_{D2}}$

**Verb post-stem suffixes**

Figure V.6 illustrates the relative ordering among verb post-stem suffixes.

| S<br><br>Verb Stem | J<br><br>Tense-Aspect Suffixes Grp-1 | K<br><br>Person Number Suffixes | L<br><br>Inter. Suffix | M<br><br>Tense-Aspect Suffixes Grp-2 | N<br><br>Person Number Suffixes |
|---|---|---|---|---|---|
| | $J_1$<br>-(E)r<br>-Hyor<br>-(y)AcAk<br>-mAlI<br>-mHş | $K_1$<br>-lAr | -mH | $M_1$<br>-(y)mHş<br>$M_2$<br>-(y)DH<br>$M_3$<br>-(y)sA | $N_1$<br>-(yH)m<br>-(y)Hz<br>-(sH)n<br>-(sH)nHz |
| | $J_2$<br>-DH<br>-sA | $K_2$<br>-(yH)m<br>-(sH)n<br>-k<br>-(sH)nHz | | $M_4$<br>-DHr | $N_2$<br>-(y)Hn<br>-(y)HnHz<br>-sHn<br>-sHnlAr |
| | $J_3$<br>-(y)A | $K_3$<br>-(yH)m<br>-lHm<br>-(sH)n<br>-(sH)nHz | | | $N_3$<br>same as $K_2$ |

Figure V.6: The relative order of verb post-stem suffixes.

Unlike the relatively simple patterns of verb stem derivation, verb post-stem suffixes exhibit complex relationships mainly because of numerous classes and subclasses with several members. The chart in Figure V.6 illustrates the relative ordering among verb post-stem suffixes. Note that suffixes and suffix alternants that fall in the same column are mutually exclusive. Orders are established in terms of frequency of occurence. Person-number suffixes that appear in two different positions belong to the same morpheme category, but have distinct ordering relations.

Let us explain the suffixes presented in the above figure. Class $J$ representing the first tense-aspect group, consists of the subclasses $J_1$, $J_2$, and $J_3$.

($J_1$) It contains a group of tense aspect suffixes.

- `-(E)r` is the aorist suffix. Its negation allomorph `-z`, occurs after the negative suffix `-mA` and can not be followed by first person singular or plural suffixes.
- `-Hyor` is the continuous aspect suffix. It represents the continuous action in or up to present time, planned future, habitual, and historic present.
- `-(y)AcAk`, `-mAlI`, and `-mHş` are future, necessitative and quoted past aspect suffixes.

($J_2$)
- `-DH` is the past tense suffix. It represents witnessed past action.
- `-sA` is the conditional aspect suffix.

($J_3$) `-(y)A` is the optative aspect suffix. It represents desire or permission.

Class $K$ and $N$ consists of person-number suffixes.

($K_1$) `-lAr` is the third person plural suffix.

($K_2$) `-(yH)m` and `-k` are the first person singular and plural suffixes and `-(sH)n` and `-(sH)nHz` are the second person singular and plural suffixes.

For ease of design, verbal post-stem suffixes are handled in parts. Considering the great number of restrictions involved, the chart given in Figure V.6 is simplified by leaving out some classes, subclasses or some of their members. The items left out are treated separately. This will leave us the core of Turkish verb inflectional system.

Let us begin with basic verb final suffixation to model verbal inflections. All the verb stems may be combined with any one of the tense-aspect suffixes listed in Figure V.6. The resulting construction can forego further combinations with the suffixes in the remaining order. Note that, the remaning suffixes (only with the exception of $N_2$ which contains the imperative person suffixes) may not occur without a preceding tense-aspect group-1 suffix. Therefore, $J$ is required as a link between $S$ and the remaining orders that follow. A simplified version of the chart in Figure V.6 is given in Figure V.7.

<table>
<tr><th>S</th><th>J</th><th>K</th><th>L</th><th>N</th></tr>
<tr><td>Verb<br>Stem</td><td>Tense<br>Aspect<br>Suffixes<br>Grp-1</td><td>Person<br>Number<br>Suffixes</td><td>Inter.<br>Suffix</td><td>Person<br>Number<br>Suffixes</td></tr>
<tr><td rowspan="2"></td><td>$J_1$<br>-Hyor<br>-(y)AcAk<br>-mAlI<br>-mHş</td><td>$K_1$<br>-lAr</td><td rowspan="2">-mH</td><td rowspan="2">$N_1$<br>-(yH)m<br>-(y)Hz<br>-(sH)n<br>-(sH)nHz</td></tr>
<tr><td>$J_2$<br>-DH<br>-sA</td><td>$K_2$<br>-(yH)m<br>-(sH)n<br>-k<br>-(sH)nHz</td></tr>
</table>

Figure V.7: A fragment of verb post-stem suffixes.

The basic difference between the subclasses $J_1$ and $J_2$ is that they take distinct person-number suffixes. Another difference in morphotactics of these classes is that the interrogative suffix `-mH` precedes[10] the person-number suffixes in the first group and follows the person-number suffixes in the latter case. Possible continuations are $J_1$ $K_1$ L and $J_1$ L $N_1$ for subclass $J_1$, and $J_2$ $K_1$ L and $J_2$ $K_2$ L for subclass $J_2$. These continuations can be formulated as

- $F_{Vi1}$= $\boxed{S\ J_1\ (k_1\ \mathrm{l} + \mathrm{l}\ N_1)}$

Note that $J_4$ and $J_5$ are introduced to denote subclasses for `-(E)r` and `-z` respectively since the negative alternant `-z` for `-(E)r` exhibits irregularities. There also exists a long-distance dependency in the morphotactics of `-z`: If the deverbal negation suffix in Class $H$ exists and the compound suffixes of Class $I$ do not exist in the sequence, `-(E)r` corresponds to the morphophonemic alternant `-z`. Therefore, `-(E)r` should be handled seperately. Besides `-z` corresponds to null when it precedes the allomorphs of the first person singular and plural suffixes `-(yH)m` and `-(y)Hz`. Examples of these occurences are presented below.

```
(55) a. gel  -mA  -z    -mH   -(yH)m  → gelmez miyim?
        come -NEG -AORS -INTR -PERS1s
        'Would I not come?'

     b. gel  -mA  -z    -(yH)m  -mH   → gelmem mi?
        come -NEG -AORS -PERS1s -INTR
        'Would I not come?'
```

Thus, possible continuation sequences for the aorist and the negative aorist are:

- $F_{Vi5a}$= $\boxed{V_R F_{D1}\ ((gHIJ_4) + J_4)\ (K_1 l + l N_1)}$

- $F_{Vi5b}$= $\boxed{V_R F_{D1}\ gH\ (J_5(LN_1 + lk_1 + N_{12}) + N_{11})}$

[10] The only exception to this is the third person plural suffix `-lAr` that makes up the subclass $K_1$ in Figure V.7.

Here, $V_R$, $N_{11}$, and $N_{12}$ denote verbal root, the subclass {`-(yH)m`, `-(y)Hz`} of $N_1$, and the subclass {`-(sH)n`, `-(sH)nHz`} of $N_1$. Thus, the formula describing the valid verbal inflections for $J_4$ and $J_5$ is as follows:

- $F_{Vi5}=\boxed{F_{Vi5a} + F_{Vi5b}}$

The post-stem penultimate verbal inflections for subclasses $J_2$ and $J_3$ have also been formulated in the same manner. These formulations are presented below. Recall that $J_2$ contains the suffixes {`-DH`, `-sA`} and $J_3$ denotes the 'optative' aspect suffix `-(y)A`.

- $F_{Vi2}=\boxed{S\ J_2\ (k_1 + K_2)\ l}$

- $F_{Vi3}=\boxed{S\ (J_3\ (k_1 + k_3)\ l) + K_4\ l)}$

($N_2$) This subclass consists of the suffixes `-(y)Hn`, `-(y)HnHz`, `-sHn`, and `-sHnlAr`. Along with the unmarked verb stem, they indicate the imperative in Turkish:

| | | | | |
|---|---|---|---|---|
| (56) | a. gel | → gel | 'come' (2nd person singular) |
| | b. gel-(y)Hn | → gelin | 'come' (2nd prs. plu. or 2nd prs. sing. formal) |
| | c. gel-(y)HnHz | → geliniz | 'come' (2nd prs. plu. or 2nd prs. sing. formal) |
| | d. gel-sHn | → gelsin | 'come' (3rd person singular) |
| | e. gel-sHnlAr | → gelsinler | 'come' (3rd person plural) |

The imperative morphotactics can be modelled as follows:

- $F_{Vi4}=\boxed{V_R\ F_{D1}\ n_2}$

Thus, the basic verb final suffixation is formulated as a combination of above formulas.

- $F_{Vi}=\boxed{F_{Vi1} + F_{Vi2} + F_{Vi3} + F_{Vi4} + F_{Vi5}}$

**Tense-aspect suffixes group-2:** The integration of Class $M$ suffixes in the system increases the complexity of our formulation greatly. There are also semantic restrictions introduced. For instance, suffix `-sA` of subclass $J_2$ may not occur with suffix `-(y)sA` of subclass $M_3$. Clearly, there is a semantic restriction that does not allow the occurence of two subsequent 'conditions' in a single sequence.

Morphotactical formulations for the tense-aspect group-2 are made according to the first tense-aspect group $J$. Note that, the formulations for subclasses $J_1$, $J_2$, $J_4$, and $J_5$ are provided here.

Note that formulations are organized for a group of $J$-$M$ couples.

- $F_{Vfi1a}$= $\boxed{J_{G1}\ (k_1lM_1 + lM_1K_1 + lM_1N_1)}$

- $F_{Vfi1b}$= $\boxed{J_{G1}\ (k_1lM_2 + lM_2k_1 + lM_2K_2)}$

- $F_{Vfi1c}$= $\boxed{J_{G2}\ (k_1lM_3 + lM_3K_1 + k_1M_3L + M_3K_1L + LM_3K_2 + M_3K_2l)}$

  where

  – $J_{G1} = (\ S\ (J_1 + J_2) + V_RF_{D1}\ (\ gHI\ J_4 + J_4 + gH\ J_5))$
  – $J_{G2} = (\ S\ (J_1 + J_2{}^{-\texttt{sA}}) + V_RF_{D1}\ (\ gHI\ J_4 + J_4 + gH\ J_5))$

Class $J_1$ tense-aspect suffixes can also combine with the copula suffix `-DHr` which introduces the emphasis or supposition on the predication. For instance,

```
(57) a. gel  -Hyor -DHr  → geliyordur
        come -CONT -COP
        ‘s/he is coming (supposedly)’

     b. var    -mHş        -lAr    -mH   -DHr → varmışlar mıdır?
        arrive -PAST(narr) -PERS3p -INTR -COP
        ‘Did they arrive (do you think)?’
```

The formula for above combinations is as follows:

- $F_{Vfi2} = J1\ (k_1 l M_4 + l M_4 K_1 + l N_1 M_4)$

The formulations for the post-stem verb-final suffixation covered here are combined as:

- $F_{Vfi}$= $\boxed{F_{Vfi1} + F_{Vfi2}}$

Eventually, the verbal paradigm covered can be represented by the formula:

- $F_V$=$\boxed{F_D + F_{Vi} + F_{Vfi}}$

# CHAPTER VI

## IMPLEMENTATION

The overview of the working system together with the peripheral systems it communicates are illustrated in Figure VI.1.

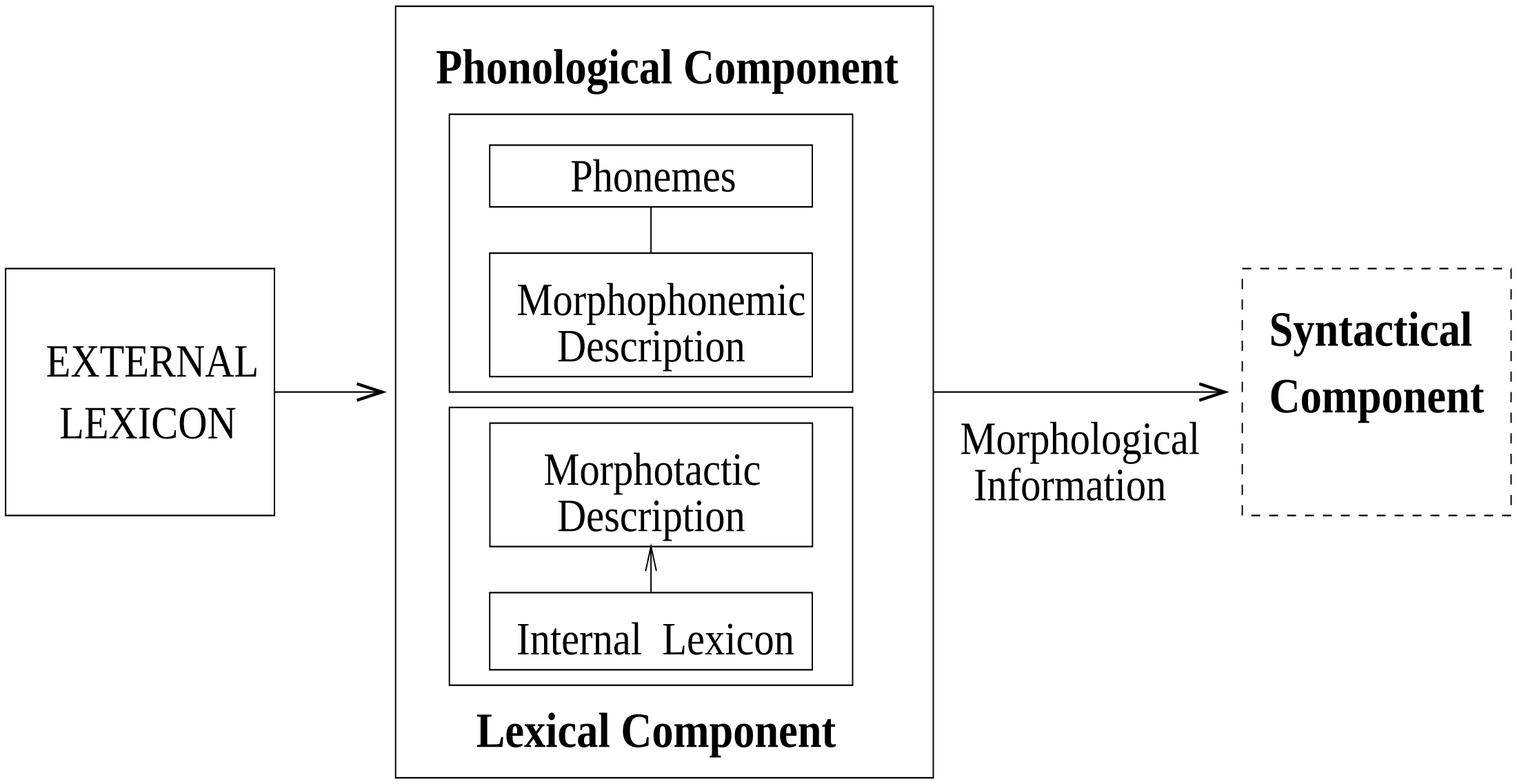

Figure VI.1: An overview of the working system

The core of the system consists of a 'rule' file and a 'lexicon' file. The two-level rules are encoded using *twolc* [37], a finite-state two-level rule compiler. *Twolc* is capable of compiling two-level rules into minimized finite-state transducers.

These transducers can also be merged into a single FST. However, as the number of rules constituting the morphophonemic description increases, the size of the merged transducer becomes very large. The size of the merged transducer is equal to the product of sizes of the individual transducers. Therefore, merging all the FSTs into one single FST should be avoided if the number of two-level rules exceeds a sensible amount, although a single FST executes faster. Our rules file consists of 49 such rules. The final finite-state transducer of the current system is a 2048 state machine consisting of 681 nets and occupying 1920 KB of memory. Meanwhile, *Twolc* together with its internal structure consumes 2712 KB of memory. When the FSTs are intersected, the number of states becomes 6357 and the number of nets decreases to 25. This time the final finite-state transducer consumes 11009 KB of memory. Relying on these observations, FSTs of the system are not merged into a single FST but left as participants of a single net. FSTs created by *twolc* can be saved in a tabular form in an ASCII file or in binary form. The binary form is preferred to the tabular form since it is considerably smaller in size. The archiphonemes used in the implementation of twol-level rules are as follows:

- `A` is the meta-phoneme that corresponds to the set of low unrounded vowels {`a, e`} according to the context. Some morphemes that have `A` are the ablative case suffix `-DAn` and the conditional tense/aspect suffix `-(y)sA`.

- `H` is the meta-phoneme that corresponds to the set of high vowels {`ı, i, u, ü`}. Some morphemes that have `H` are the first person singular and plural suffixes `-(H)m` and `-(H)mHz`, respectively.

- `C` is the suffix-initial meta-phoneme that corresponds to the set of consonants {`c, ç`}. It also exists as the lexeme final stop in some monosyllabic roots such as `ölC` 'measure'.

- `D` is the suffix initial meta-phoneme with possible correspondences of {`d, t, /null/`}. Some morphemes that have `D` are the locative case suffix `-DA` and the causative suffix `-(D)HX`. Note that `D` is also used as the lexeme final meta-phoneme to designate the context where the lexeme final `d` is not voiced (e.g., `aD` 'name').

- `B` is used as the lexeme final consonant that corresponds to `b`. It designates the context where the lexeme final consonant `b` is not devoiced (e.g., `aB` 'water').

- `K` is used as the lexeme final consonant that corresponds to `k` that does not become voiced (e.g., `göK` 'sky').

- `q` is used as the lexeme final consonant of the disyllabic loaned words such as `hukuq` 'law'. It corresponds to a `k` that does not become voiced.

- `R`, `U`, and `P` are utilized as the meta-phonemes of the artificial prefix `RUP-` which is used to model reduplication processes.

- `Y` is the meta-phoneme that is used to model the `suY` 'water' exception. It corresponds to the set {`y, /null/`} according to the context.

- `N` corresponds to the pronomial `n` or `/null/` according to the context.

- `L` is used as the morpheme initial meta-phoneme of the third person plural

possesive suffix `-LArH`. It corresponds to the set {`l, /null/`} according to the context and designates the context where the segment `LAr` drops after a preceding number suffix `-lAr`.

- `X` represents the morpheme final meta-phoneme of the causative suffixes `-(DH)X` and `-(D)HX`. It corresponds to the set {`t`, `r`} depending on the context.

The finite-state lexicon compiler *lexc* [25] contains the morphotactic description. The lexical items consist of free and bound forms. *Lexc* requires all the lexical items together with their initial continuation class, since it constructs a tree-like structure representing the finite-state machine at run time in memory where free forms are the entry points of the structure. *Lexc* as well as *twolc* minimize the finite-state machinery they construct. *Lexc* composes the FST net produced by *twolc* and the FSA net created by itself into a single net for a complete description of the language in use. This final net is also saved in a binary form to be processed by an external tool.

For this work, we have constructed a tool called *trk* with a moderate graphical user interface which provides an integrated user-friendly environment. The main unit of this tool is a **C** program that consists of an engine that reads in the complete language description encoded in the specified binary format, and analyze/generate word forms using the language-specific description. The library functions to parse and load binary description file were provided by Xerox. We developed the program according to our needs and embodied it in the integrated environment. *Trk* has been equipped with extra facilities which allows users to change and recompile the rules file, to rebuild the final language description net, and edit/print the lexicon files via a graphical user interface. The integrated environment is to be developed in accordance with the changing needs in future. A view of *Trk* is given in Figure VI.2.

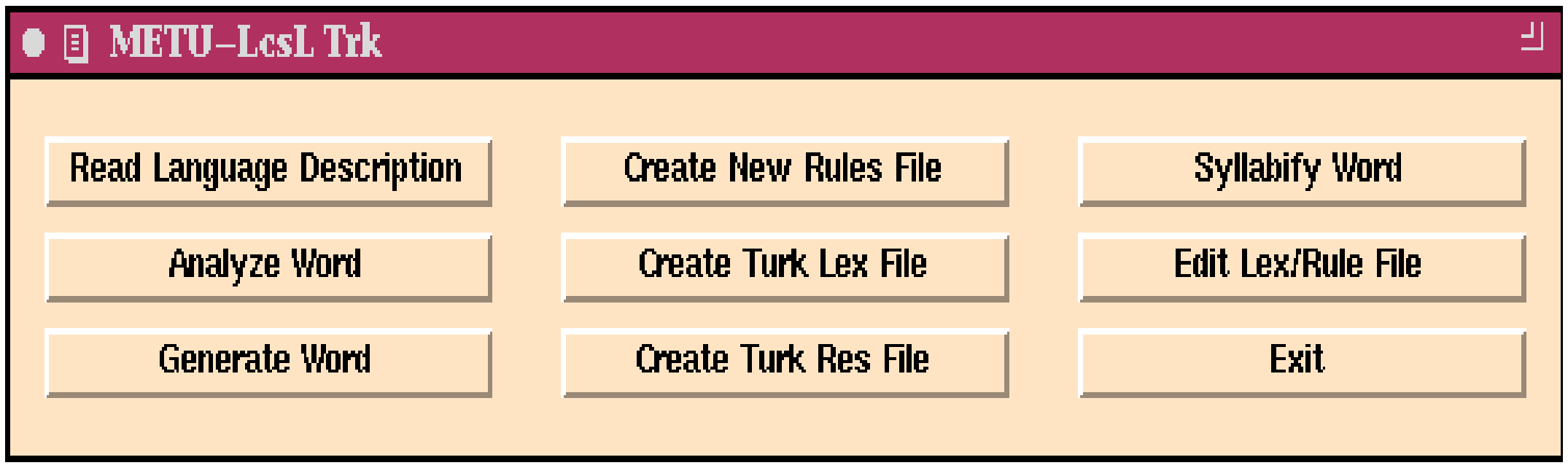

Figure VI.2: The user-interface of the system

## VI.1 Integration with peripheral components

*Trk* is not meant to be a stand-alone system. The interfacing mechanisms that link the system to the peripheral components such as the external lexicon and the syntactic parser should be defined. Firstly, let us briefly discuss the the relation between the lexicon and the morphological analyzer. The external lexicon that is designed to fulfill the requirements of other natural language processing modules such as syntactic parser and semantic analyzer provide much more information than that required by the morphological analyzer. The mechanism to fetch required information from the external lexicon should be developed. For the time being, the lexical entries together with their continuation classes are kept in the internal lexicons. These are text files constructed according to the syntax described by *lexc*. Fragments of lexicon files for nominal and verbal roots are given in Figure VI.3 and Figure VI.4 respectively.

```
[ROOT=uç]:uç^  V0_CONT;
[ROOT=var]:vaR^ V0_CONT;

LEXICON V1
[ROOT=bak]:bak^ V1_CONT;
[ROOT=bek]:bek^le V1_CONT;
[ROOT=bil]:biL^ V1_CONT;
[ROOT=böl]:böL^ V1_CONT;
[ROOT=boya]:bo^ya  V1_CONT;
[ROOT=bul]:buL^ V1_CONT;
[ROOT=dè]:dè^ V1_CONT;
[ROOT=din]:din^le V1_CONT;
```

Figure VI.3: A fragment of the internal verbal lexicon.

```
 [ROOT=gazete]:ga^zete NC_CONT;
 [ROOT=giysi]:giy^si NC_CONT;
 [ROOT=gök]:göK^ NC_CONT;
 [ROOT=hac]:hacc^ NC_CONT;
 [ROOT=hak]:hakk^ NC_CONT;
 [ROOT=harf]:hárf^ NC_CONT;
 [ROOT=harb]:hárb^ NC_CONT;
 [ROOT=hat]:hatt^ NC_CONT;
 [ROOT=hukuk]:hu^kuq NC_CONT;
 [ROOT=elbise]:el^bise NC_CONT;
 [ROOT=usul]:usúl^ NC_CONT;
 [ROOT=tuz]:tuz^ NC_CONT;
 [ROOT=vakit]:vá^kHt NC_CONT;
 [ROOT=zabıt]:za^bHt NC_CONT;

LEXICON NC1
 [ROOT=ağız]:a^ğHz NC1_CONT;
 [ROOT=alın]:a^lHn NC1_CONT;
 [ROOT=ayak]:a^yak NC1_CONT;
 [ROOT=bacak]:ba^cak NC1_CONT;
```

Figure VI.4: A fragment of the internal lexicon for noun roots.

We have to point out the fact that the lexical items should have been entered in the external lexicon according to the specifications outlined in the design chapter. That is, lexical representations should follow the norms so that two-level rules can use them. For instance, the noun root 'book' should be represented as `kitab` but not `kitap`. The word forms extracted from the external lexicon are appended to the morphotactic description making up the lexicon file. Another requirement for the free forms is that they should have been syllabified, since syllabification provides critical information for most of the morphophonemic rules. Especially the lexeme-final stop voicing/devoicing rules use the first syllable marker to designate the proper context for application. We also provided a utility program to syllabify a list of word forms with *Trk*.

## VI.2 Future Improvements

One of the major problems with the finite-state model is that at any state in the finite-state machine, the transition to be made is decided depending upon the current state. Under such a design constraint, it is very hard to encode the morphotactics for languages such as Turkish which have complex morphologies. This also introduces a considerable amount of redundancy in morphotactical desription. Turning back to the formulas presented in Chapter V, recall that many alternant sequences (i.e., "+") exist. For instance, let us examine the formula of the verbal paradigm.

- $F_{Vi2a}$= $\boxed{SF_{D1}(((GH+H)IJ_4)+J_4)(K_1l+lN_1)}$

Note that a Class $I$ suffix is forced to precede Class $J_4$ suffix `-(E)r`, if the negation suffix `-mA` has been utilized before. In doing so, the subsequences $GHI$ and $HI$ disallow illegal parsers as in example.

```
(58) a. gel  -mA  -(E)r -mH   -(yH)m  → *gelmermiyim
        come -NEG -AORS -INTR -PERS1s
        ‘ ’
```

Let us visualize that $H$ and/or $I$ are supposed to be optional. Then invalid instances like `*gelmer`, `*geleyebilir` are possible. Although the relative ordering is correct, occurrences related with long distance dependency can not be resolved. Dependency relation among two nonconsecutive suffixes can only be established by duplicating the path between two suffixation points. Looking from the other side, to allow a valid combination like `gelmez` and to disallow an invalid one like `*gelmeyebilz`, a formulation that involves a permanent $H$ adjacent to $J_5$ has to be introduced. Note that this sequence totally discards $I$, i.e., even an optional $I$ is not permitted. The second formulation is then as follows:

- $F_{Vi2b}$=$\boxed{SF_{D1}gHJ_5lN_1}$

It is obvious that the second formula creates redundancy, since most of the states that are used in the first formulation are duplicated. Yet, there is this no way around in the finite-state domain.

This problem is easily overcome in a context-free domain. The information of relative ordering among affixes is adequte to constitute a word grammar for a context-free parser. The critical issue here is the assignment of correct features to morphemes making up the word grammar. A context-free parser with a feature-unification engine performs morphological analyis. Therefore, a contribution to this study can be made by the integration of context-free parsing mechanism like that of PC-KIMMO-2 with the current finite-state structure for handling the morphotactics. Let us mention briefly about the new features of PC-KIMMO-2 and its advantages before stating how the current model relying on a finite-state structure can be ported to a context-free structure.

PC-KIMMO-2 introduces major improvements to the original two-level framework. One of these new features, the word grammar, is the most significant. The word grammar component uses a unification-based parser. It consists of context-free rules and feature constraints. One obvious difference between parsing a sentence and parsing a word is that a sentence is typically already tokenized into words while a word is not tokenized into morphemes. In PC-KIMMO-2, the recognizer uses the rules and the lexicon to tokenize a word into a sequence of morphemes which in turn is passed to the word grammar component for parsing. There are several advantages of such an approach.

First of all, the word grammar component offers a more powerful model of morphotactics. PC-KIMMO-1 used only the continuation class model of morphotactics which was used in Koskenniemi's original model [19, 20]. *Lexc* [25] of Xerox which is used to model Turkish morphotactical paradigm also utilizes the continuation class model. In continuation class model, the morphotactic properties of a morpheme can be stated only in terms of the classes of morphemes that can directly follow it in a word. This means that it is very difficult or at least practically infeasible to enforce certain discontinuous dependencies between morphemes. The word grammar, however, has the full power of a context-free grammar and can model the word structure as arbitrarily complex branching trees. The practical result is that with PC-KIMMO-2, one can eliminate most of the bad parses that were so difficult to eliminate in version 1.

Secondly, the word grammar component can deduce the lexical category (part-of-speech) of a word. PC-KIMMO-1 and *Lexc* could break a word into its morphemes and gloss each morpheme, but it could not tell the category of the whole word. For example, given the word `çözümleme`, the finite-state lexicon compilers would return the analysis as

```
(59) a. çöz   -(H)m  -lA    -mA
        solve -NVDER -VNDER -NVDER
        'the act of solving'
```

The original word has been broken into four morphemes with glosses, but there is no indication that the whole word is a noun. This deficiency made the finite-state tools less usable as a front-end to a syntactic parser, since a syntactic parser must know the category of each word. In PC-KIMMO-2, the feature passing mechanism can be used to determine the lexical category of a word.

# CHAPTER VII

## CONCLUSION

In this chapter, we discuss the results and the problems experienced during the study, and suggest some possible extensions to the current work.

The distinctive features of our model can be summarized as follows:

- Finite-state machines are used to model morphotactics.
- Finite-state transducers are used to model morphophonemics.
- Complete word grammars for nominal and verbal paradigms are described.
- Loan-word phonology, as well as native phonology, is taken into account. For clarity, they are kept as separate but related models.
- Morpheme representations, including bound and free forms, have explicit notational devices such as archiphonemes, optional segments and syllable boundaries.

A comprehensive morphological analysis is a necessity for agglutinative languages like Turkish if one requires the information related with word structures;

syntax is one place where this information is crucial. In Turkish, beginning with a word root, both inflectional and derivational suffixes are stacked together into one string, the translation of which into a non-agglutinative language requires a multitude of words. This stacking process sometimes exhibits recursive structure as in the case of relativization, which makes it all the more infeasible to put all inflections and derivations in the huge lexicon. It is obviously inefficient to put all the word forms below generated from the word root `house` in the lexicon.

(60)

| | | | |
|---|---|---|---|
| 1. | evde | ev-DA | house-LOC |
| 2. | evdeki | ev-DA-ki | house-LOC-REL |
| 3. | evdekiler | ev-DA-ki-lAr | house-LOC-REL-PLU |
| 4. | evdekilerin | ev-DA-ki-lAr-(n)Hn | house-LOC-REL-PLU-GEN |
| 5. | evdekilerinki | ev-DA-ki-lAr-(n)Hn-ki | house-LOC-REL-PLU-GEN-REL |

Secondly both phonology and morphotactics can be implemented using a finite-state approach. However such a finite-state approach to morphotactics causes many problems both in design and implementation if long-distance dependencies exist in the word grammar. For instance, noun bases that take predicative verb prefinal suffixes for the present aspect undergo such a long distance dependency. Given the example below, note that the information that the noun stem is plural has to be carried along up to the predicative person-number suffix to prevent illegal parses. The only way around to preserve this information in a finite-state model is to duplicate the states and transitions in between.

(61) *kardeşlerimsin brother-PLU-POSS1s-PERS2s *'you are my brothers'

Although these long distance dependencies can be solved by duplicating certain transitions, the redundancy introduced diminishes the expressive power and generic representation of the word grammar.

The two-level model has its strength in its phonological component. Morphophonemics can effectively be modelled and implemented using finite-state machinery. Tools of two-level phonology provide tractable end-products that can handle the morphophonemic processes as an integral part of a morphological analyzer/generator. For instance, *twolc* converts the two-level rules to deterministic, minimized finite-state transducers both in tabular and binary formats that can be easily processed. Our model consists of 52 morphophonemic rules and 258 possible continuation classes for morphotactics. *Twolc* compressed 2048 state machines to 1242 state machines for FST processing. Average CPU times for some examples are[1] shown in Table VII.1.

Table VII.1: The processing times for sample executions.

| 3 morpheme words | 6 morpheme words |
|---|---|
| 1250 words/sec. | 671 words/sec. |

Currently, our lexicon consists of 73 bound forms (i.e., suffixes) and 182 free forms (i.e., words) of different types.

As stated above, the morphotactic component of the two-level model is relatively weak, whereas the phonological component is very powerful. One potential improvement for morphotactics is to use a context-free grammar and unification. A unification-based parser working on a word grammar proposed recently in PC-KIMMO-2 seems to be an appropriate and effective mechanism for handling morphotactics. Such a unification-based approach eliminates the redundancies introduced by long-distance dependencies, and provides a better interface both to the lexical and the syntactic component. Recall that a lexical component con-taining all the roots and affixes together with the unification features is required for the word grammar. Actually this coincides with the claim of Rochelle Lieber [23] that there is no separate component of morphology in the grammar; lexical, morphological, and syntactical information are parts of the whole.

[1] CPU times are for a Sun Sparcstation 10.

# REFERENCES

[1] Antworth, Evan L. 1990. *PC-KIMMO: a two-level processor for morphological analysis*. Occasional Publications in Academic Computing No. 16. Dallas, TEXAS:Summer Institute of Linguistics.

[2] Banguoğlu, Tahsin. 1990. *Türkçe'nin Grameri*. Ankara: Türk Tarih Kurumu Basım Evi.

[3] Barton, G. Edward. 1986. Computational complexity in two-level morphology. In *ACL Proceedings, 24th Annual Meeting* (Association for Computational Linguistics).

[4] Barton G. Edward., Robert C. Berwick., and Eric S. Ristad. 1987. *Computational complexity and natural language*. Cambridge, MA: The MIT Press.

[5] Bear, J. 1986. A morphological recognizer with syntactic and morphological rules. In *COLING-86* (Association for Computational Linguistics).

[6] Beesley, R. Kenneth. Computer analysis of Arabic morphology:A two-level approach with detours. In *Proceedings of the Third Annual Symposium on Arabic Linguistics* (University of Utah).

[7] Bozkurt, Fuat. 1992. *Türklerin Dili*. İstanbul : Cem Yayınevi.

[8] Can, Kaya. 1986. *Üniversite ve Yüksekokullar için Türk Dili* (1.Kitap). ıstanbul: Kıral Matbaası.

[9] Chomsky, Noam. 1965. On certain formal properties of grammars. In R.D. Luce, R.R.Bush, and E.Galenter, eds., *Readings in mathematical psychology*, Vol. II, 323-418. New York: John Wiley and Sons.

[10] Chomsky, N. and Halle, M. 1968. *The Sound Pattern of English*. New York: Harper and Row. (This is normally cited as SPE.)

[11] Hankamer, J. 1986. Finite-state morphology and left to right phonology. In *Proceedings of the West Coast Conference on Formal Linguistics*, Volume 5.

[12] Hopcroft, John E., and Jeffrey D. Ullman. 1979. *Introduction to automata theory, languages, and computation*. Reading, MA: Addison-Wesley Publishing Company.

[13] Holman, E. 1988. FINNMORF: A computerized reference tool for students of Finnish morphology. *Computers and the Humanities*, 22:165-172.

[14] Jakobson, R., Fant, G. and Halle, M. 1952. *Preliminaries to speech analysis: the distinctive fatures and their correlates.* MIT Acoustics Laboratory. 5th printing 1963. Cambridge, Mass.: MIT Press.

[15] Johnson, C. Douglas. 1972. *Formal aspects of phonological description.* The Hague: Mouton.

[16] Katamba, Francis. 1993. *An introduction to phonology.* London: Logman Group UK Limited.

[17] Kay, Martin. 1983. When meta-rules are not meta-rules. In Karen Sparck Jones and Yorick Wilks, eds., *Automatic natural language parsing*, 94-116. Chichester:Ellis Horwood Ltd.

[18] Kay, Martin. 1987. Non-concatenative finite-state morphology. In *ACL Proceedings, 24th Annual Meeting* (Association for Computational Linguistics).

[19] Koskenniemi, Kimmo. 1983. Two-level model of morphological analysis. In *IJCAI-83* (International Joint Conference on Artificial Intelligence).

[20] Koskenniemi, Kimmo. 1984. A general computational model for word-form recognition and production. In *COLING-84* (Association for Computational Linguistics).

[21] Koskenniemi Kimmo., and Kenneth W. Church. 1988. Complexity, two-level morphology and Finnish. In *Proceedings of Coling '88*, 335-340. (Association for Computational Linguistics).

[22] Klavans, J., and Chodorow, M. 1988. Using a mor phological analyzer to teach theoretical morphology. Technical report RC 13794, IBM Thomas J. Watson Research Center.

[23] Lieber, Rochelle. 1992. Deconstructing Morphology. The University of Chicago Press.

[24] Lees, B. R. 1961. The Phonology of Modern Standard Turkish. *Research and Studies in uralic and Altaic Languages*, Project 13. Indiana University, Bloomingtoon.

[25] Karttunen, Lauri. 1993. *Finite-State Lexicon Compiler.* Copyright 1993 Xerox Corporation.

[26] Allen, J., Hunnicutt, M. S., and Klatt, D. 1987. *From Text to Speech: The MITalk System.* Cambridge University Press.

[27] Solak, Ayşin and Oflazer, Kemal. 1993. Design and implementation of a spelling checker for Turkish. *Literary and Lnguistic Computing*, Vol. 8, No:3.

[28] Oflazer, Kemal. 1994. Two-level description of Turkish morphology. *Literary and Linguistic Computing*, Vol. 9, No:2.

[29] Sebüktekin, Hikmet, I. 1974. Morphotactics of Tukish Verb Suffixation. *Boğaziçi Üniversitesi Dergisi.* pp:87-117.

[30] Van Schaaik, G., J. 1995. Studies in Turkish Grammar. Ph.D. Dissertation. Universiteit van Amsterdam.

[31] Shieber, Stuart M. 2-7 July 1984. The Design of a Computer Language for Linguistic Information. In *Proceedings of Coling '84 10th International Conference on Computational Linguistics*, 362-366. Stanford University, Stanford, California.

[32] Shieber, Stuart M. 1986. An Introduction to Unification Based Approaches to Grammar. *CSLI Lecture Notes Series*, Number 4, Center for the Study of Language and Information, Stanford University.

[33] Sproat, Richard William. 1992. *Morphology and Computation.* The MIT Press.

[34] *Türkçe Büyük Sözlük.* 1980. Türk Dil Kurumu.

[35] Tekin, Talat. 1995. Türkçe'de morfofonemik degişmeler: Dar ünlü nöbetleşmesi. *Dilbilim Araştırmaları.* Ankara: Bizim Büro Basımevi.

[36] Göçmen, Elvan., Şehitoğlu, Onur., and Bozşahin, Cem. 1995. An outline of Turkish Syntax. Technical Report TR 95-2 METU Computer Engineering Department. Ankara.

[37] Karttunen, Lauri. and Beesley, R. Kenneth. 1992. ' *Two-Level Rule Compiler.* Copyright 1992 Xerox Corporation.

[38] Underhill, Robert. 1976. *Turkish Grammar.* MIT Press.

# APPENDIX A

## THE TWO-LEVEL RULES FILE

```
ALPHABET
 %' %-:0 %+:0 %*:0 a b c ç d e f g ğ h ı i j k l L:l m n o ö p r s ş t u ü v
 y z D:d %(:0 %):0 # P:p P:m P:s P:r C:ç Y:0 N:0 K:k è:e á:a ó:o ú:u I: E:
 X:r B:b q:k R:r U: P: ;

SETS
 Cs = b c ç d f g ğ h j k l m n p r s ş t v y z ;
 CsVed = c ğ d b ;
 CsVless = ç f g h k p s ş t ;
 CsLatNas = l r m n v ;

 Ccp = g ;
 Cğp = s ;
 Clp = d y ;
 Cnp = c y ;
 Crp = d k s ;
 Cvp = s ;
 Cyp = k ;
 Czp = k t ;

 Cds = b ;
 Cğs = d ;
 Cls = b ;
 Cms1 = p ;
 Cms2 = t ;
 Cps = t ;
 Crs = m ;
 Cts = b k ;
 Cvs = m y ;
 Cys = b ;

 Ccm = s ;
 Ckm = b d s ;
 Crm = b ;
 Csm = y ;
```

```
 Cşm = b y ;
 Cym = b s;
 Czm = d ;

 Cbr = ç ;
 Cmr = t ;

 CBS = p ; !  bilabial voiceless stop p
 CBN = m ; !  bilabial [voiced] nasal m
 CAF = s ; !  alveolar voiceless fricative s
 CAL = r ; !  alveolar [voiced] non-lateral liquid r

 V = a e ı i u ü o ö á è ó ú ;
 Vfr = i e ü ö á è ó ú ;
 Vfrurd = i e è ;
 Vfrrd = ü ö ó ú ;
 Vbk = a ı u o ;
 Vac = á ó ú ;
 Vacrd = ó ú ;
 Vbkurd = a ı ;
 Vbkrd = o u ;
 Nnhgrd = o ö ;
 SIC = n s y ; !  Suffix-initial Consonant Head
 SIV = A H E ; !  Suffix-initial Vowel Head


DEFINITIONS
 RB = %+:0 ; !  Root Boundary
 MB = %-:0 ; !  Morpheme Boundary
 FSM = %^:0 ; !  First syllable marker
 OPHL = %(:0 ; !  Optional suffix head's left paranthesis
 OPHR = %):0 ; !  Optional suffix head's rigth paranthesis
 OPAP = %' ; !  Optional apostrophe that is used to signal proper nouns.

!Some composite definitions
 MJ = [ OPAP | RB | FSM | MB]* ;
!MJ is the Morphme Juncture.
!Morpheme Juncture consists of zero or more repetition of the markers that
!can exist at the morpheme boundary..No ordering relation what so ever, has
!been imposed on them.

 OPVD = OPHL (D:0) V:0 OPHR;
!The context where optional vowel drops:
 ! a  r  a  b  a  -  (  H  )  m   or   a  n  l  a  -  (  D  H  )  X
 ! a  r  a  b  a  0  0  0  0  m        a  n  l  a  0  0  0  0  0  t

 OPCD = OPHL SIC:0 OPHR;
!The context where suffix-initial consonant drops:
 ! e  v  -  (  s  )  H  N
 ! e  v  0  0  0  0  i  0

 OPC = OPHL [ D: (V:) | SIC: ] OPHR ;
!The context where optional suffix-initial consonant exists.
!Don't care if it drops or not:
```

```
 ! e  v  -  (  s  )  H  N   or   a  r  a  b  a  -  (  s  )  H  N
 ! e  v  0  0  0  0  i  0        a  r  a  b  a  0  0  s  0  ı  0

 OPV = OPHL SIV: OPHR ;
!ev00i0m or araba0000m

 OPNC = OPHL :Cs OPHR ;
!The context when optional consonant drop does not occur.  For instance:
 ! k  a  p  ı  -  (  s  )  H  N  -  D  A
 ! k  a  p  ı  0  0  s  0  ı  n  0  d  a

 OPNV = OPHL :V OPHR ;
!The context when optional suffix-initial vowel does not drop:
 ! e  v  -  (  H  )  m
 ! e  v  0  0  i  0  m

 OPCV = OPHL { D:0 | SIC:0 } H: OPHR;  OPD = { OPC | OPV };
!The context of possible suffix initial drop of consonant and vowel couple.

!The contexts for vowel harmony rules.

LSV = ((OPHR)(Cs:)(:Cs)(OPCV)([(Y:)|(N:)](FSM))(RB)(OPAP)|OPNC|:Cs*V:0|OPCV);
!The Optional Left Side of MB for Vowel Harmony

LSVH = [((LSV) MB (OPV | OPCV) :Cs*) | LSV | FSM Cs* H:0 Cs* ];
!The Left-hand Side of Morpheme Boundary for Vowel Harmony rules.

RSVH = { :Cs* | OPC :Cs* | OPCV :Cs* | OPHL | OPHL D: | L:0 A:0 r:0 };
!The Right-hand Side of Morpheme Boundary for Vowel Harmony rules.

!LCVH = LSVH (FSM) (RB) (MB) RSVH ;
!LCVH2 = :Cs* [[:0 - FSM]* | [L:l | R:r] FSM (:0)*];

SLCVH = [:Cs* (V:0) :Cs* (’) (:0)*]*;
!Simplified Left Context for Vowel Harmony.

RCFSDV = (FSM) (’:)  (RB) [ # | MB (:0*) [OPNC | (OPC) :Cs]];
!The right context for Final Stop Devoicing.


RULES
"1.Vowel Harmony, A:a " A:a => [:Vbk - Vac:] SLCVH _ ;
!A isn’t always realized as a (A:0 is also possible) in the above environment
!but this environment is the only environment in which A:a is allowed.

"2.Vowel Harmony, A:e "
A:e => [:Vfr | Vac:] SLCVH _ ;

"3.Vowel Harmony, H:u "
H:u => [:Vbkrd - Vacrd:] SLCVH _ ;

"4.Vowel Harmony, H:ü "
H:ü => [:Vfrrd | Vacrd:] SLCVH _ ;


"5.Vowel Harmony, H:ı "
H:ı => [:Vbkurd - á:a] SLCVH _ ;
```

```
"6.Vowel Harmony, H:i "
H:i => [:Vfrurd | á:a] SLCVH _ ;

"7.Vowel Harmony, I:ı "
I:ı => :Vbk SLCVH _ ;

"8.Vowel Harmony, I:i "
I:i => :Vfr SLCVH _ ;

!  Rules for AORIST suffix
"9.Vowel Harmony for AORIST tense"
E:a <=> :Vbk [:Cs+ - [ L:l | R:r]] FSM (RB) MB OPHL _;

"10.Vowel Harmony for AORIST tense"
E:e <=> :Vfr [:Cs+ - [ L:l | R:r]] FSM (RB) MB OPHL _;

"11.Vowel Harmony for AORIST tense"
E:u => [:Vbkrd - Vacrd:] SLCVH _ ;

"12.Vowel Harmony for AORIST tense"
E:ü => [:Vfrrd | Vacrd:] SLCVH _ ;

"13.Vowel Harmony for AORIST tense"
E:ı => [:Vbkurd - á:a] SLCVH _ ;

"14.Vowel Harmony for AORIST tense"
E:i => [:Vfrurd | á:a] SLCVH _ ;

"15.Final Stop Devoicing"
CsV:CsVless <=> \[CsV:] _ (CsV:0 - c:0) RCFSDV;
where
   CsV in (b c d)
   CsVless in (p ç t)
matched;

"16.Stop Voicing ç:c "
ç:c <=> _ (:0 - FSM)* MB (:0)* :V;
       :V CsLatNas _ (:0 - FSM)* (FSM) (RB) MB (:0)* :V;

"17.Stop Voicing k:ğ "
k:ğ <=> :V _ (:0 - FSM)* MB (:0)* :V;

"18.Stop Voicing g:ğ "
g:ğ <=> o _ (:0 - FSM)* MB (:0)* :V;

"19.Stop Voicing k:g "
k:g <=> :V n _ (:0)* :V;
!I exclude the special correspondence t:d for the final final stop t since as
!Gerdines Johannes van Schaaik claims "Only 690 out of 1835 nouns ending in t
!are subject to that mapping.  That is 37 percent".

"20.Suffix-Initial Devoicing"
D:t <=> :CsVless (:0)* MB _ ;

"21.Suffix-Initial Voicing"
C:c <=> \:CsVless (:0)* MB _ ;
```

```
"22.SIC-DELETION (n,s,y):0"
SIC:0 <=> :Cs+ (:0  [N:0 | Y:0 ])* MB OPHL _ OPHR;
where SIC in (n s y);

"23.SIV-DELETION (H,A,E):0"
SIV:0 <=> :V (:0-[Y: | N:])* (FSM) (RB) MB OPHL _ OPHR;
where SIV in (A H E);

"24.The rule for Continuous aspect suffix '-Hyor'."
!It is one of the most crosswise suffix of Turkish.  If there is a vowel at
!the word boundary, H replaces with that vowel.  Note that it is valid for all
!vowels but è.
Vow:0 <=> _ [(:0)* - Cs:0] (RB) MB H: y o r;
where Vow in (a e ı i u ü o ö á ó ú);

"25.The rule for è:i correspondence."
!The high e (i.e., è) in the monosyllabic verbal roots dè and yè becomes
!i when verbal suffixes with an initial (y) segment immediately follows.
!These suffixes are -(y)AcAk and -(y)A."
è:i <=> {d | y} _ (FSM) (RB) MB [ H: y o r | OPHL y OPHR];

"26.Correspondence for the archiphoneme R of Reduplication"
R:Csx <=> _ :V P: MB Csy ;
where
   Csx in (b c ç d g k m p s t y)
   Csy in (b c ç d g k m p s t y)
matched;

"27.Correspondence for the archiphoneme U (i.e., the vowel)of Reduplication"
U:Vx <=> :Cs _ P: MB :Cs Vy ;
       R:0 _ P: MB Vy ;
where
   Vx in (a e ı i u ü o ö)
   Vy in (a e ı i u ü o ö)
matched;

"27.b.  Disallow R:0 in other contexts.."
R:0 /<= _ (FSM) MB;

!Rules for REDUPLICATION
"28.Correspondences for the archiphoneme P of Reduplication"
!P is always realized as one of the consonants m p r s
!according to the mappings given but this is not the only environment
!that this realization occur.
!CBS : bilabial voiceless stop p.
!CBN : bilabial [voiced] nasal m.
!CAF : alveolar voiceless fricative s.
!CAL : alveolar [voiced] non-lateral liquid r.
!CRD : Reduplication of Consonant.
P:CBS <=> :Cs U: _ MB Cx Vz Cz ;
        R:0 U: _ MB V Cs ;
where
   Cx in (Ccp Cğp Clp Cnp Crp Cvp Cyp Czp)
   Cz in (c ğ l n r v y z)
   Vz in (V Vfr V V V V V V)
matched;
```

```
"29.  " P:CBN <=> :Cs U: _ MB Cx Vz Cz ;
where
   Cx in (Ccm Ckm Crm Csm Cşm Cym Czm)
   Cz in (C k r s ş y z)
   Vz in (V V V V V Vfr V) matched;

"30."
P:CAF <=> :Cs U: _ MB Cx Vz Cy ;
where
   Cx in (Cds Cğs Cls Cms1 Cms2 Cps Crs Cts Cvs Cys Czs)
   Cy in (d ğ l m m p r t v y z)
   Vz in (V V V V Vbk V V V V Vbk V)
matched;

"31.Reduplication of the dental-alveolar non-lateral liquid r"
P:CAL <=> :Cs U: _ MB Cx Vz Cy ;
where
   Cx in (Cbr Cmr Cpr)
   Cy in (b m p)
   Vz in (V Vfr V)
matched;

"32.Degemination"
Cy:0 <=> [Cx:  - Cx:0] _ (FSM) (RB) [ # | MB OPHL y:0 OPHR :Cs | MB :Cs];
where
   Cx in (b c d f h k l m n r s t v y z)
   Cy in (b c d f h k l m n r s t v y z)
matched;

"33.High vowel epenthesis in word bases"
H:0 <=> :V FSM Cs _ :Cs (RB) MB [OPCD :V | OPNV | :V];

"34.Instantiation of the pronominal n (N:n)"
N:n <=> _ (RB) MB [OPHL y:  OPHR [H: \z | A:] | D: A: (:n) ];
        _ FSM (RB) MB OPNV;

"35.The rule for su [suY] exception."
Y:y <=> _ (FSM) (RB) MB [OPNV | :V | OPC];

"36.The rule for ‘bana’ and ‘sana’ exceptions."
e:a <=> [b|s] _ n FSM (RB) MB OPHL y:  OPHR A: [# | MB];

"37.The two-level rule representing the epenthesis of the initial phoneme
‘H’ of the ‘-Hyor’ morpheme drops after a lexeme final vowel."
!Note that this ‘H’ is not optional.  H replaces with the vowel of the
!preceding lexeme, unless the preceding lexeme is a monosyllabic verbal root.
!There are two such verbal roots namely ‘dè’ and ‘yè’.  In this case, è
!replaces with ‘H’. That is ‘H’ is let to drop.
H:0 <=> [d | y] è:i FSM (RB) MB _ y o r;
        [:V | :l | :r] (RB) MB OPHL D:0 _ OPHR X: ;
        [:V | :l | :r] (RB) MB OPHL D:0 OPHR _ X:t ; "38.D:t"
D:t <=> [ \r k | [:CsVless-:k] ] (:0)* MB OPHL _ H: OPHR X:;
        [[:CsVless-[:ç|:ş|:t|:k]|[\:r :k]]FSM|:CsVless](RB)MB OPHL _ OPHR H: X:;
```

```
"39.The rule for various causative suffixes."
D:0 <=> [V|l|:r|:Vrk FSM] (RB) MB OPHL _ H: OPHR X:;

[:V|l|:r|[[:ç|:ş|:t|:r:k]FSM]](RB) MB OPHL _ OPHR H: X:;

"40.The rule for the causative suffixes."
X:t <=> [V|l|:r|:V rk FSM] (RB) MB OPHL [D:0 H: OPHR | D:0 OPHR H:] _ ;

"41.The rule for ‘-LArHN’."
L:0 <=> MB l A: r MB _ ;

"42.The rule for ‘-LArHN’."
L:l /<= MB l A: r MB _ ;

"43.The rule for ‘-LArHN’."
A:0 <=> MB L:0 _ ;

"44.The rule for ‘-LArHN’."
r:0 <=> MB L:0 A:0 _ ;

"45.The rule for passive voice suffix."
l:n <=> [:V | l] (RB) MB OPHL H:0 OPHR _ ;

"46.The rule for Negative Aorist suffix:  ‘z:0’."
!It maps to NULL when it precedes the first person singular or plural
!person-number suffixes.
z:0 <=> MB m A: MB _ MB OPHL y:  ;

"47.The rule for allomorphic variations of -(yH)m."
y:0 <=> z:0 MB OPHL _ H: OPHR m;
         :Cs MB OPHL _ H: OPHR;
        MB ( OPHL y:  OPHR ) [ D: H: | s:  A: ] MB OPHL _ H: OPHR ;

"48.The rule for the allomorphic variations of ‘-(sH)n’ and ‘-(sH)nHz’:‘s:0’"
s:0 <=> MB ( OPHL y:  OPHR ) [ D: H: | s:  A: ] MB OPHL _ H: OPHR;

"49.The rule for the allomorphic variations of ‘-(sH)n’ and ‘-(sH)nHz’:‘H:0’"
H:0 <=> z:0 MB OPHL y:0 _ OPHR m;
       MB ( OPHL y:  OPHR ) [ D: H: | s:  A: ] MB OPHL [y:0 | s:0] _ OPHR ;

"50.The rule for passive voice where -(H)L has the allomorphs -(H)l, -(H)n"
L:n <=> [:V | :l] (FSM) (RB) (MB) OPHL H: OPHR _ ;

"51.The rule for the case in which the word final apostrophe corresponds to 0"
’:0 <=> _ #;

"52.The rule for A:0 the preceding the ‘-Hyor’ suffix."
A:0 <=> m _ MB H: y o r;
```

# APPENDIX B

## SPECIAL CORRESPONDENCES

Table B.1: Special correspondences for meta-phonemes.

| LR | SR | | | | | | |
|---|---|---|---|---|---|---|---|
| A | a | e | (null) | | | | |
| H | ı | i | u | ü | (null) | | |
| I | ı | i | u | ü | | | |
| E | a | e | ı | i | u | ü | (null) |
| P | m | p | r | s | | | |
| L | l | n | (null) | | | | |
| Y | y | (null) | | | | | |
| N | n | (null) | | | | | |
| D | d | t | (null) | | | | |
| C | c | ç | | | | | |
| R | Cs | r | (null) | | | | |
| U | Vow | | | | | | |
| X | r | t | | | | | |
| B | b | | | | | | |
| K | k | | | | | | |
| e | e | a | | | | | |
| á | a | | | | | | |
| é | e | i | | | | | |
| ó | o | | | | | | |
| ú | u | | | | | | |
| b | b | p | | | | | |
| c | c | ç | | | | | |
| d | d | t | | | | | |
| g | g | ğ | | | | | |
| k | k | ğ | | | | | |
| q | k | | | | | | |
| n | n | (null) | | | | | |
| r | r | (null) | | | | | |
| s | s | (null) | | | | | |
| y | y | (null) | | | | | |
| z | z | (null) | | | | | |